%% file: main.tex
\newif\ifshowcomments
\showcommentstrue
\ifshowcomments
\newcommand{\mynote}[2]{\fbox{\bfseries\sffamily{#1}}
 {\small$\blacktriangleright$\textsf{#2}$\blacktriangleleft$}}
\else
\newcommand{\mynote}[2]{}
\fi

\documentclass[sigconf]{acmart}
\AtBeginDocument{%
  }

\setcopyright{acmlicensed}
\copyrightyear{2025}
\acmYear{2025}
\acmDOI{XXXXXXX.XXXXXXX}
\acmConference[Submitted to KDD 2025]{Make sure to enter the correct
  conference title from your rights confirmation emai}{August 03--07,
  2025}{Toronto, ON, Canada}
  
\acmISBN{978-1-4503-XXXX-X/18/06}

\usepackage{booktabs}
\usepackage{tabularx}
\usepackage{caption}
\usepackage{graphicx}
\usepackage{enumitem}
\usepackage{float}
\begin{document}
\setlength{\parindent}{0pt}

\title{Beyond Black-Box Benchmarking: Observability, Analytics, and Optimization of Agentic Systems}

\author{Dany Moshkovich\textsuperscript{\textdagger}}
\affiliation{%
  \institution{IBM Research - Israel}
  \city{Haifa}
  \country{Israel}}
\email{mdany@il.ibm.com}

\author{Hadar Mulian\textsuperscript{\textdagger}}
\affiliation{%
  \institution{IBM Research - Israel}
  \city{Haifa}
  \country{Israel}}
\email{Hadar.Mulian@ibm.com}

\author{Sergey Zeltyn\textsuperscript{\textdagger}}
\affiliation{%
  \institution{IBM Research - Israel}
  \city{Haifa}
  \country{Israel}}
\email{sergeyz@il.ibm.com}

\author{Natti Eder}
\affiliation{%
  \institution{IBM Research - Israel}
  \city{Haifa}
  \country{Israel}}
\email{netanele@il.ibm.com}

\author{Inna Skarbovsky}
\affiliation{%
  \institution{IBM Research - Israel}
  \city{Haifa}
  \country{Israel}}
\email{inna@il.ibm.com}

\author{Roy Abitbol}
\affiliation{%
  \institution{IBM Research - Israel}
  \city{Haifa}
  \country{Israel}}
\email{roy.abitbol@il.ibm.com}

\begin{abstract}

The rise of agentic AI systems, where agents collaborate to perform diverse tasks, poses new challenges with observing, analyzing and optimizing their behavior. Traditional evaluation and benchmarking approaches struggle to handle the non-deterministic, context-sensitive, and dynamic nature of these systems.
This paper explores key challenges and opportunities in analyzing and optimizing agentic systems across development, testing, and maintenance. We explore critical issues such as natural language variability and unpredictable execution flows, which hinder predictability and control, demanding adaptive strategies to manage input variability and evolving behaviors. Through our user study, we supported these hypotheses. In particular, we showed a 79\% agreement that non deterministic flow of agentic systems acts as a major challenge. Finally, we validated our statements empirically advocating the need for moving beyond classical benchmarking. 
To bridge these gaps, we introduce taxonomies to present expected analytics outcomes and the ways to collect them by extending standard observability frameworks.
Building on these foundations, we introduce and demonstrate novel approach for benchmarking of agent evaluation systems.  Unlike traditional “black box” performance evaluation approaches, our benchmark is built from agent runtime logs as input, and analytics outcome including discovered flows and issues. By addressing key limitations in existing methodologies, we aim to set the stage for more advanced and holistic evaluation strategies, which could foster the development of adaptive, interpretable, and robust agentic AI systems.

\end{abstract}


\begin{teaserfigure}
  \includegraphics[width=\textwidth]{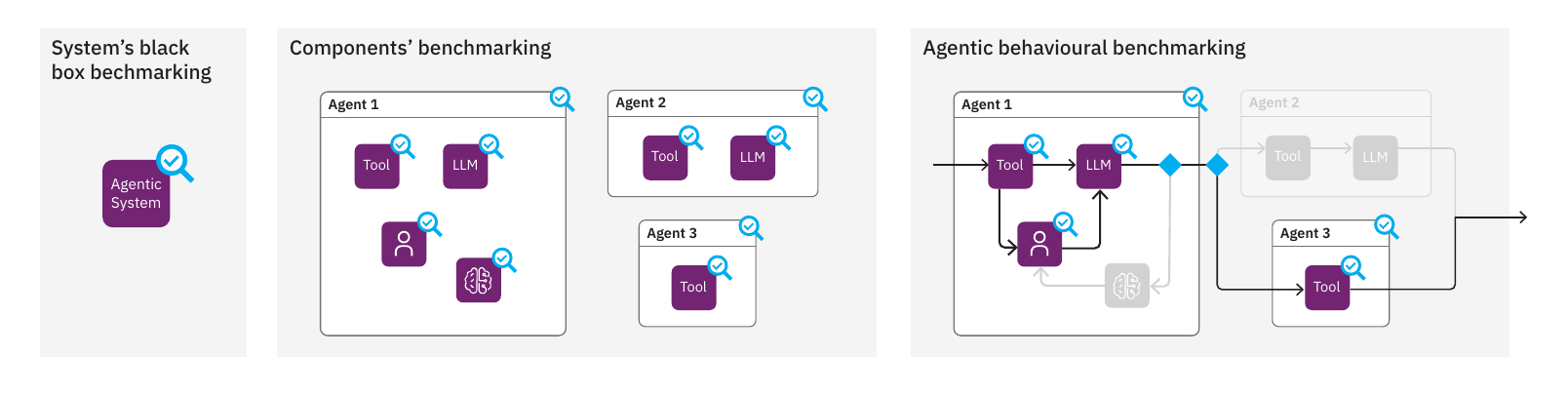}
  \caption{Beyond Outcomes: Understanding Agentic Behavior}
  \Description{Our teaser figure}
  \label{fig:main}
\end{teaserfigure}

\maketitle

\section{Introduction}

\label{sec:introduction}

\input{Sections/introduction}

\section{Related Art}

\label{sec:related_art}

\input{Sections/related_work}

\section{Agentic System Analysis: Gaps and Requirements}

\label{sec:requirements}

\input{Sections/requirements}

\section{Agent Observability and Analytics Approach}

\label{sec:taxonomies}

\input{Sections/taxonomy}

\section{Benchmarking Agent Analytics Technologies }

\label{sec:benchmarking}

\input{Sections/benchmarking}

\section{Discussion and Future Work}

\label{sec:future_work}

\input{Sections/future_work}

\bibliographystyle{ACM-Reference-Format}
\bibliography{agent_analytics}

\appendix

\input{Sections/appendix}

\end{document}

%% file: Sections/introduction.tex
The rapid evolution of Large Language Models (LLMs) has given rise to agentic AI systems—complex architectures where multiple agents collaborate to perform diverse tasks. While these systems unlock unprecedented capabilities in problem-solving and task automation, they also introduce new challenges in observability, analysis, and optimization, requiring approaches that go beyond traditional benchmarking.

Unlike conventional software with deterministic behavior patterns, agentic systems operate within dynamic and evolving environments, where their decision-making processes and outputs can exhibit variability. This variability may arise from several factors, including the stochastic nature of LLMs, the interdependencies among system components, and the impact of language-based interactions. Even minor differences in input phrasing or task conditions could influence execution flows and final responses \cite{he2024does,sun2024investigation,salinas2024butterfly,loya2023exploring}. 

These challenges highlight the limitations of conventional evaluation methods, particularly black-box benchmarking, which assesses system performance without capturing reasoning processes and interactions. Addressing issues like unpredictable execution flows, complex task decomposition, and multi-framework interactions requires new evaluation methodologies.

We highlight the need for Agentic System Behavioral Benchmarking, as illustrated in Figure \ref{fig:main}, to go beyond outcome-based metrics by analyzing execution patterns, decision-making, and interactions, while capturing non-deterministic flows, response variability, and system dynamics.



These new approaches will require enhanced observability, analysis, and optimization by incorporating both structured and natural language inputs while accounting for inherent uncertainty.



Our work takes initial steps in these directions and presents several key contributions:
\begin{enumerate}[left=0pt]
    \item We identify critical gaps in existing agentic system evaluation methods, particularly the limitations of traditional benchmarking in capturing non-deterministic behavior and performance, emphasizing the need for a shift to behavioral benchmarking.  

    
    \item We validate these gaps through empirical experiments and a user study, highlighting their impact across development, testing, and maintenance.  
    \item We present Agentic System Behavioral Benchmarking, defining core agentic system elements and introducing semantic conventions for observability and analytics taxonomies.
    %
    %
    \item We propose a novel benchmarking methodology for evaluating \textbf{agent analytics} technologies, encouraging them to assess agentic system behavior beyond classical benchmarking, and present ABBench (agent Analytics Behavioral Benchmark) dataset that follows this methodology. 
\end{enumerate}




The remainder of this paper is organized as follows: Section \ref{sec:related_art} reviews related work in monitoring and analytics of agentic systems. Section \ref{sec:requirements} identifies key gaps in the evaluation of agentic systems.
Section \ref{sec:taxonomies} defines new taxonomies for observability and performance analytics, providing a foundation for systematic evaluation. Section \ref{sec:benchmarking} presents a benchmarking framework tailored for agentic systems, alongside a dedicated dataset designed to enhance agent analytics and evaluation. 
Finally, Section \ref{sec:future_work} concludes with a discussion of implications and future work.

%% file: Sections/related_work.tex
The evolution of software system monitoring parallels the growing complexity of applications. Initially, debugging and monitoring relied on unstructured logs, providing only rudimentary insights into system behavior. As software architectures became more distributed and complex, the need for standardized monitoring and analytics methods emerged. \textbf{OpenTelemetry (OTel)} \cite{blanco2023practical}  represents a significant advancement in establishing such standards, integrating logs, traces, and metrics into a unified observability framework.

The rise of interactive software systems, particularly web applications, introduced new monitoring challenges. These distributed systems required enhanced approaches to capture and analyze both system performance and user behavior\cite{jansen2022understanding, zheng2015web}.

Even before the advent of large language models (LLMs), AI-driven software systems demanded sophisticated analytics due to their inherent non-deterministic nature. Early conversational AI systems (chatbots) highlighted unique challenges in monitoring and analytics, requiring solutions that could handle large volumes of interaction data while providing meaningful insights into conversation patterns and system behavior \cite{Yaeli21}.

These early challenges laid the foundation for modern approaches to AI system traceability.


\label{subsec:related_art_existing_technologies}

The emergence of LLM-based applications has spawned specialized observability tools that extend traditional concepts to address unique challenges, particularly their non-deterministic nature and complex interaction patterns.

\textbf{OpenLLMetry},  implemented by Traceloop \cite{openllmetry}, extends OTel standards to capture LLM-specific metrics such as token usage, cost, and latency. The framework provides comprehensive support for LLM frameworks and vector databases, enabling end-to-end observability of complex LLM applications.
\textbf{LangSmith} \cite{langsmith},  approaches observability from an LLM application lifecycle perspective, offering capabilities for debugging, evaluation, and testing. It introduces the concept of "runs" (analogous to OTel “spans”) and utilize "traces" concepts to capture the complete journey of LLM interactions, with distinctive evaluation capabilities supporting model assessment and scoring. 
Similar tools include \textbf{LangTrace} \cite{langtrace}, \textbf{LangFuse} \cite{langfuse} and others that are beyind the scope of this paper.

Comparative analysis of these frameworks \cite{jose2024harnessing} reveals varying approaches to fundamental observability challenges. While sharing core features, the tools differ in their integration capabilities, security features, and specialized functionalities. Open-source solutions offer greater flexibility but demand more technical expertise, while some tools prioritize standardization through OTel adoption and others focus on specialized LLM evaluation and debugging features. 


\label{subsec:related_art_benchmarking}

Benchmarking approaches as evaluation for agentic systems encompass both descriptive and comparative evaluations. Key evaluation dimensions include quantitative metrics (execution time, resource utilization, success rates) and qualitative aspects (decision quality, behavioral consistency). Modern frameworks emphasize multi-faceted evaluation scenarios while incorporating real-world deployment considerations such as scalability and regulatory compliance\cite{zoller2004benchmarking}.

%% file: Sections/requirements.tex
\subsection{Unique Characteristics of Agentic Systems}
Agentic systems leveraging large language models (LLMs) present unique challenges due to their inherent variability in execution and decision-making. Unlike traditional software systems with deterministic behavior, LLM-based systems generate responses probabilistically, making their outputs sensitive to input phrasing, model state, and sampling strategies. These challenges manifest in two key ways:

\begin{itemize}[left=0pt]
    \item \textbf{Flow Variability:} Agentic systems exhibit significant divergence in execution paths despite being provided with identical inputs. This variability arises due to stochastic decision-making and dynamic state dependencies within multi-agent interactions.
    \item \textbf{Natural Language (NL) Variability:} Variability in natural language inputs affects execution outcomes, as differences in phrasing or contextual framing can lead to divergent model interpretations. Additionally, even when given identical inputs, the system may produce different outputs across different runs due to inherent randomness in LLM generation.    
\end{itemize}

To quantitatively demonstrate these issues, we designed experiments to test their occurrence and measure their extent across different settings.


\subsection{Illustrative Example: A Calculator Agentic Systems}

\label{subsec:calculator}

To ground our discussion, we use an agentic calculator system that processes mathematical expressions and natural language inputs, such as  
\(\big((6 + 2) \times [8 - 3 \times 2]\big) + \{\text{Average of 3, 7, and five?}\}\).  
It handles nested expressions, diverse bracket types, and both parallel (running in different threads) and distributed (executing across multiple servers) computation, demonstrating non-deterministic responses, multi-agent interactions, dynamic flow generation, and integration with multiple frameworks and LLMs, while remaining accessible for illustrating core concepts.

The calculator decomposes expressions into sub-expressions, and a LangGraph-based \cite{langgraph} execution planner ensures correct operation sequencing and dependency management, iteratively validating results. If resolution fails, the system returns \texttt{None}.

\begin{figure}[h]
    \centering
    \includegraphics[scale=0.3]{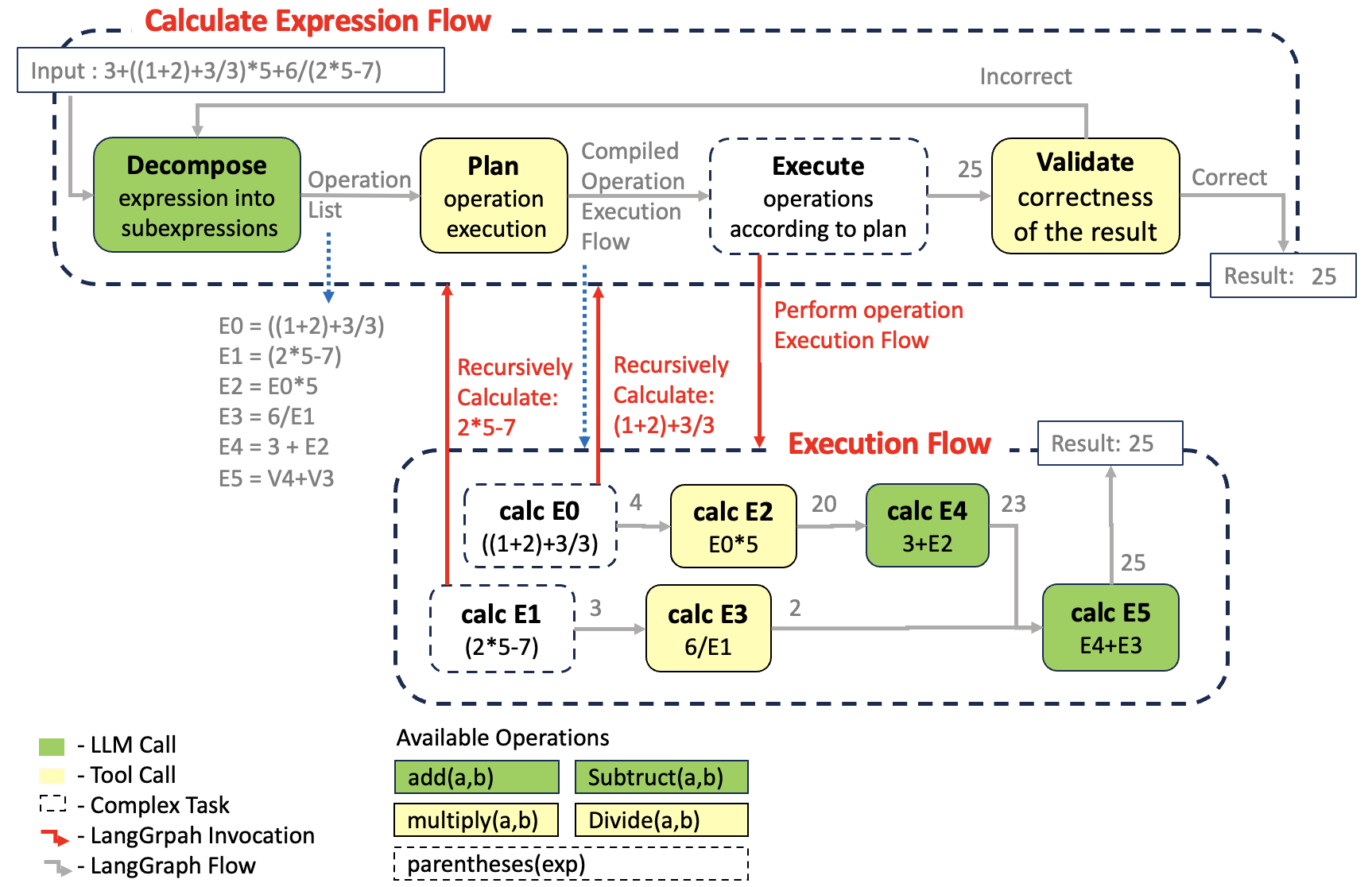} 
    \Description{A diagram presenting a calculation flow of aan expression showing multiple steps of agent, both statically defined and dynamically created.}
    \caption{Example of a calculator flow}
    \label{fig:example}
\end{figure}
For implementation details, please refer to our open-source project\footnote{\url{https://github.com/genai-analytics/publications}}.

\subsection{Experimentation Setup and Results}

\label{subsec:experiment_setup}

Based on the design and functionality of the calculator, we conducted an experiment in which the system was evaluated on 50 examples. Each example was processed five times, and the results across all repetitions were recorded.

To assess performance accuracy, we computed the mean squared error (MSE). Variability between multiple executions of the same input was quantified using the mean graph-edit distance (GED) between graphs of the execution flow generated from each repeated run of the agentic system, as follows: 
\begin{align*} \label{ged_score}   
 \frac{1}{\binom{n}{2}} \sum_{1 \leq i < j \leq n} \text{GED}(G_i, G_j)
\end{align*}
Where $n$ stands for the total amount of runs and was set to 5. Cost was measured in US dollars, based on OpenAI-4o API usage, with the number of tokens tracked per execution. Execution time was recorded in seconds. For each of these dimensions, the coefficient of variation (CV) for each example was calculated as a measure of variability within examples. 

To assess the impact of LLM-induced variations, the experiment was conducted across two example variations, with the mean coefficient of variation computed for each dimension as a summarized variability score. More details on the datasets can be found in Section \ref{sec:appendix_datasets} of the Appendix.

\subsubsection{Flow Variability:}
Execution flow differences were primarily assessed using edit distances between execution traces for identical inputs. As shown in Table 1, results for the "pure math" dataset, which consists of standard mathematical expressions, exhibit a mean CV of $63\%$ across all examples. This indicates substantial variability in both the number of tasks executed by the agentic system and the overall structure of the execution paths. The high mean percentages observed across other tracked performance dimensions further reinforce the presence of significant divergence in execution behavior.

\subsubsection{NL Variability:}
To assess the impact of natural language (NL) interpretation, the experiment was conducted on a dataset comprising of mathematical problems written in natural language. Results indicate a mean CV of $19\%$ in accuracy, demonstrating that identical inputs can yield inconsistent outputs across multiple runs. This confirms that ambiguity in NL inputs contributes to interpretational variability, which in turn affects overall system performance.
These findings align with the accuracy results obtained for the "pure math" dataset, where clearly structured mathematical expressions were consistently interpreted by the system, yielding stable outputs across repetitions. 

While absolute values for other performance dimensions, such as the average number of LLM calls per example, were elevated for NL inputs, the mean CV percentages remained consistent to those of the "pure math" dataset.

\begin{table}[h!] 
    \centering
    \small
    \setlength{\tabcolsep}{4pt} 
    \captionsetup{aboveskip=0pt, belowskip=0pt}
    \begin{tabular}{lccccc}
        \toprule
        Dataset & Accuracy & Cost & Execution Time & LLM Calls & Variability \\
        \midrule
        Pure Math & 0\% & 43\% & 45\% & 45\% & 63\% \\
        Pure NL & 19\% & 42\% & 46\% & 43\% & 64\% \\
        \bottomrule
    \end{tabular}
    \caption{Coefficient of variation across performance dimensions.}
    \label{tab:results}
\end{table}

\subsection{User Study}
\label{subsec:user_study}
To further understand these challenges, we conducted a user study with 38 practitioners. Findings highlight non-deterministic execution as a major issue (80$\%$), with key pain points including agents not following instructions (66$\%$) and inappropriate tool selection (63$\%$). Many practitioners find existing tools insufficient, with 60$\%$ reporting that current analytics do not meet their needs and 77$\%$ struggling with root cause diagnosis. Despite diverse workflows, 76$\%$ prioritize agentic flow understanding as their top analytics need. These results underscore the urgent need for improved debugging and observability in agentic systems. Additional details on the user study are available in Section \ref{sec:detailed_user_study} of the Appendix.

\subsection{Limitations and the Need for a Comprehensive Agentic System Approach}
\label{subsec:limitations}
The experimental results suggest that LLM-driven agentic systems may exhibit non-determinism in both execution flow and output, complicating performance evaluation and system reliability. This variability stems not only from the stochastic nature of LLMs but also from deeper challenges, such as:

\begin{itemize}[left=0pt]
    \item \textbf{Context Sensitivity:} System behavior can shift based on subtle prompt variations or preceding interactions.
    \item \textbf{Interactive and Temporal Dependencies:} Execution is influenced by prior steps and cumulative reasoning, making outcomes unpredictable over extended interactions.
    \item \textbf{Multi-Agent Dependencies:} In collaborative multi-agent settings, individual agent behaviors are interdependent, amplifying variability and introducing intricate coordination challenges.
\end{itemize}
Moreover, technical hurdles such as trace complexity, data scalability, and explainability further hinder effective observability. 

These challenges underscore the limitations of traditional evaluation methodologies, which struggle to capture the dynamic and evolving nature of LLM-based systems.

\textbf{The Need for a New Approach:} Given these complexities, black-box benchmarking alone is insufficient to evaluate agentic systems effectively. While benchmarking remains a valuable tool, it must be supplemented with agent analytics frameworks that provide deeper insights into execution patterns, variability, and failure modes.

To address these limitations, we propose a comprehensive agent analytics approach that integrates enhanced observability, adaptive evaluation techniques, and robust metrics and optimizations tailored for multi-agent LLM-based systems. 
This approach, alongside a newly developed benchmark for agent analytics, will be further detailed in the following sections.

%% file: Sections/taxonomy.tex

\subsection{Towards Behavioral Benchmarking}



A natural extension to traditional benchmarking involves applying it recursively to sub-components, which provides deeper insights but is still inadequate for evaluating the previously discussed unique characteristics of complex agentic systems. We propose extending benchmarking to analyze both the behaviors and interactions among components, assessing not only outcomes but also the processes by which they are achieved. Figure \ref{fig:main} illustrates this approach.

This necessitates a standardized framework for consistent observability data collection across diverse agentic systems and a set of analytical methods to leverage this data for evaluation and optimization. The following sections will define the Core Entities of AI Agentic Systems, outline the Observability of Agentic Systems, and introduce an Agentic System Analytics Taxonomy for structured monitoring, analysis, and optimization.



\subsection{Core Entities of AI Agentic Systems}
To understand our approach, we first need to examine the key entities of agentic systems, focusing on their non-deterministic properties that we aim to observe, analyze, and optimize.

An agentic system consists of interconnected components that enable intelligent, autonomous operations. Resources provide inputs, constants, and outputs, while tools are modular units that agents invoke or generate dynamically. Workflows coordinate execution by structuring tasks, which define discrete units of work handled by tools. Agents leverage resources, tools, and workflows to perform tasks, adapting through memory and learning. Organizations extend this by defining roles, enabling multi-agent collaboration.

\begin{figure}[H]
    \centering
    \includegraphics[scale=0.4]{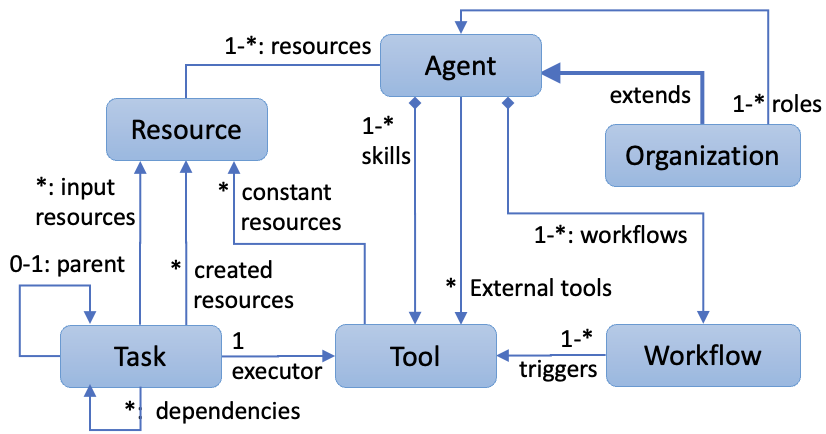} 
    \caption{Agentic System Entity Diagram}
    \Description{A diagram showing the entities of an agentic system an their relationships.}
    \label{fig:agentic_entity}
\end{figure}

Figure \ref{fig:agentic_entity} presents the entity relationship diagram, and the following sections provide an overview of each element. A more detailed description of these entities can be found in our open-source project\footnote{\url{https://github.com/genai-analytics/publications}}.

\begin{itemize}[left=0pt] 
    \item \textbf{Resources} are fundamental to an agentic system, serving as inputs, constants, and outputs that drive execution. They encompass various types, including text, images, audio, video, code, templates (often used for prompt definitions), and databases.
    \item \textbf{Tools} are basic programmatic units that can be operated or generated by the agent. Unlike classical function calls, tools reflect the non-deterministic nature of agentic systems, where their input, method of invocation, selection, or generated outcomes are inherently unpredictable. Tools can be categorized based on their non-deterministic functionality, including those for ML or LLM model inference, vector database tools for semantic search, human-in-the-loop tools for user interactions, and more complex tools for executing multi-step workflows.Tools can be wrapped with guardrails to enforce constraints, validate I/O, and guide behavior within set boundaries.
    \item \textbf{Workflows} Workflows define a structured graph of operations, coordinating tool execution and managing flow. Unlike rigid processes, agentic workflows handle non-determinism and can be dynamically generated. They incorporate various control flow relationships, such as sequencing, conditionals, joins, and forks. Some frameworks, like LangGraph\cite{langgraph}, natively support this structure for flexible execution.
    \item \textbf{Tasks} are discrete units of work executed by tools, serving as the foundation of workflows. Each task operates on input resources and data, producing resources and output data. As seen in our calculator example, a task can be decomposed by an agent into sub-tasks, which are executed by multiple tools within predefined or dynamically generated workflows. Tasks may have dependencies, ensuring proper execution order, and can be dynamically generated based on system needs. They can also be scheduled with start and end times and prioritized to optimize execution flow. While user-level tasks are first-class entities in frameworks like CrewAI \cite{crewai}, we propose extending this concept to all agentic tool invocations for a more unified and complete execution model.
    \item \textbf{Agents} are typically autonomous entities that perform tasks by leveraging their skills, implemented as tools, and triggering associated workflows. They utilize available resources and can operate external tools they have access to, ensuring effective execution according to provided instructions. Agents may follow a basic ReAct \cite{yao2023react} pattern or operate within predefined or dynamically generated workflows, as seen in our calculator example. They are often stateful entities with short- and long-term memory, enabling them to learn and adapt over time.
    \item \textbf{Multi-agent Organizations} represent structured groups of agents collaborating to achieve complex objectives. They extend agentic capabilities by defining roles, responsibilities, and constraints. Similar to Mixture of Experts (MoE) \cite{jiang2024mixtral}, they enable efficient task distribution, coordination, and specialization, leveraging diverse skills and resources.
    Frameworks like CrewAI incorporate similar concepts, represented as Crew.
\end{itemize}

\subsection{Observability of Agentic Systems}

\label{subsec:observability_taxonomy}

Observability in Agentic Systems is still an emerging field \cite{dong2024taxonomy}. While OTeL project \cite{blanco2023practical} has begun incorporating concepts related to GenAI \cite{robbins2024otel} its taxonomy remains largely focused on LLM calls, lacking comprehensive support for Agentic Systems.

In OTeL, spans represent individual operations, forming the building blocks of traces. Parent-child relationships define dependencies, while traces capture the full request lifecycle as a hierarchical structure. Context propagation ensures continuity across services.

This aims to extend OTeL’s semantic conventions for GenAI Agentic Systems by introducing GenAI Events, which bind to spans to capture the lifecycle and state changes of agentic entities in a standardized manner. These events will track key transitions, including creation, update, start, end, suspension, abortion, failure, and deletion. Collecting these events will enable analytics to monitor and optimize system behavior over time.

Each event may include a list of affected entities with their fields or identifiers, along with associated issues categorized by severity: critical error, warning, info, or debug. A single span may have multiple GenAI Events, providing a structured and standardized way to stream relevant agentic information and associated issues. This approach supports manual instrumentation by developers, ensuring deeper insights into agentic system behavior.

\subsection{Agentic System Analytics Taxonomy}
\label{subsec:analytics_taxonomy}

Analytics in Agentic Systems can be categorized into three main types: Passive, Exploratory, and Interventional, each offering different levels of insight, control, and optimization.

\begin{figure}[H]
    \centering
    \includegraphics[width=0.5 \textwidth]{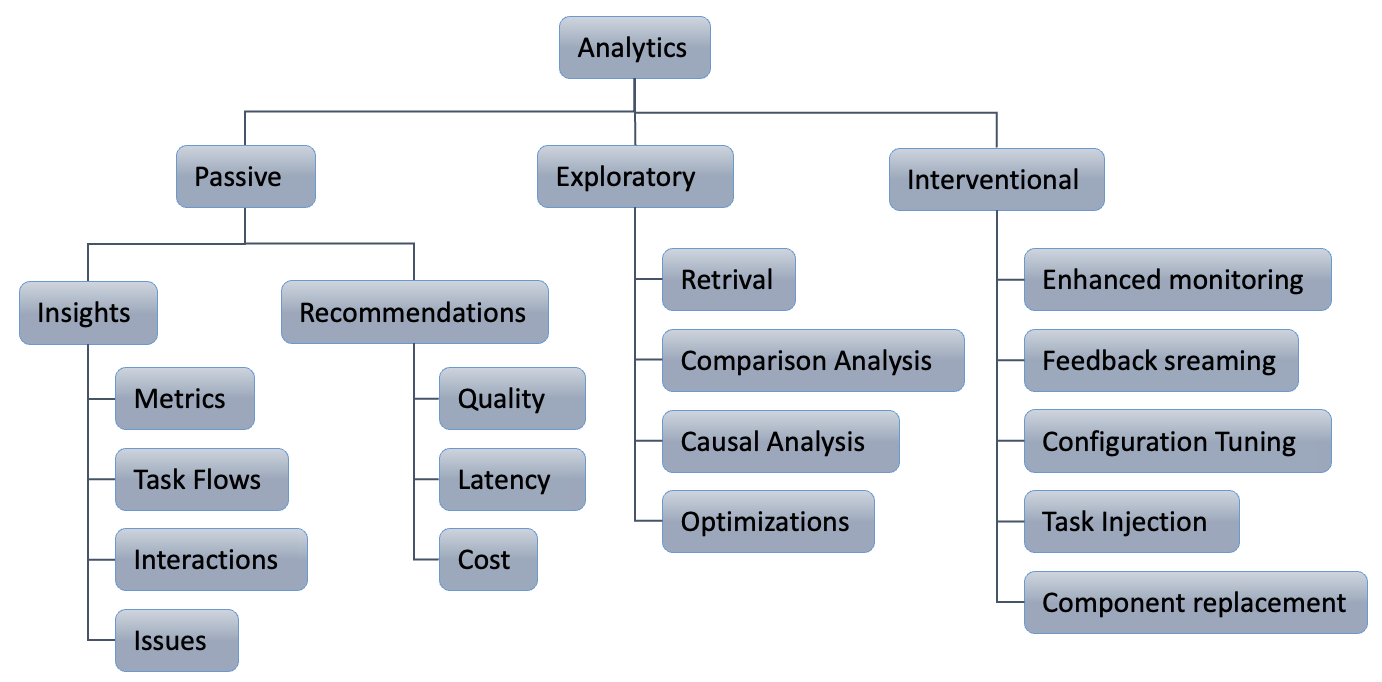} 
    \caption{Analytics Taxonomy}
    \Description{A diagram showing the taxonomy of agent analytics.}
    \label{fig:analytics_taxonomy}
\end{figure}

\textbf{Passive analytics} extracts structured insights and generates recommendations without direct user intervention, triggered periodically or upon specific events.

\begin{itemize}[left=0pt]
    \item \textbf{Insights} at both the task and workflow trace levels, including cost, latency, token usage and many others.
    \item \textbf{Recommendations}: focus on quality improvements, latency reduction, and cost efficiency based on observed patterns.
\end{itemize}

\begin{figure}[H]
    \centering
    \includegraphics[width=0.5 \textwidth]{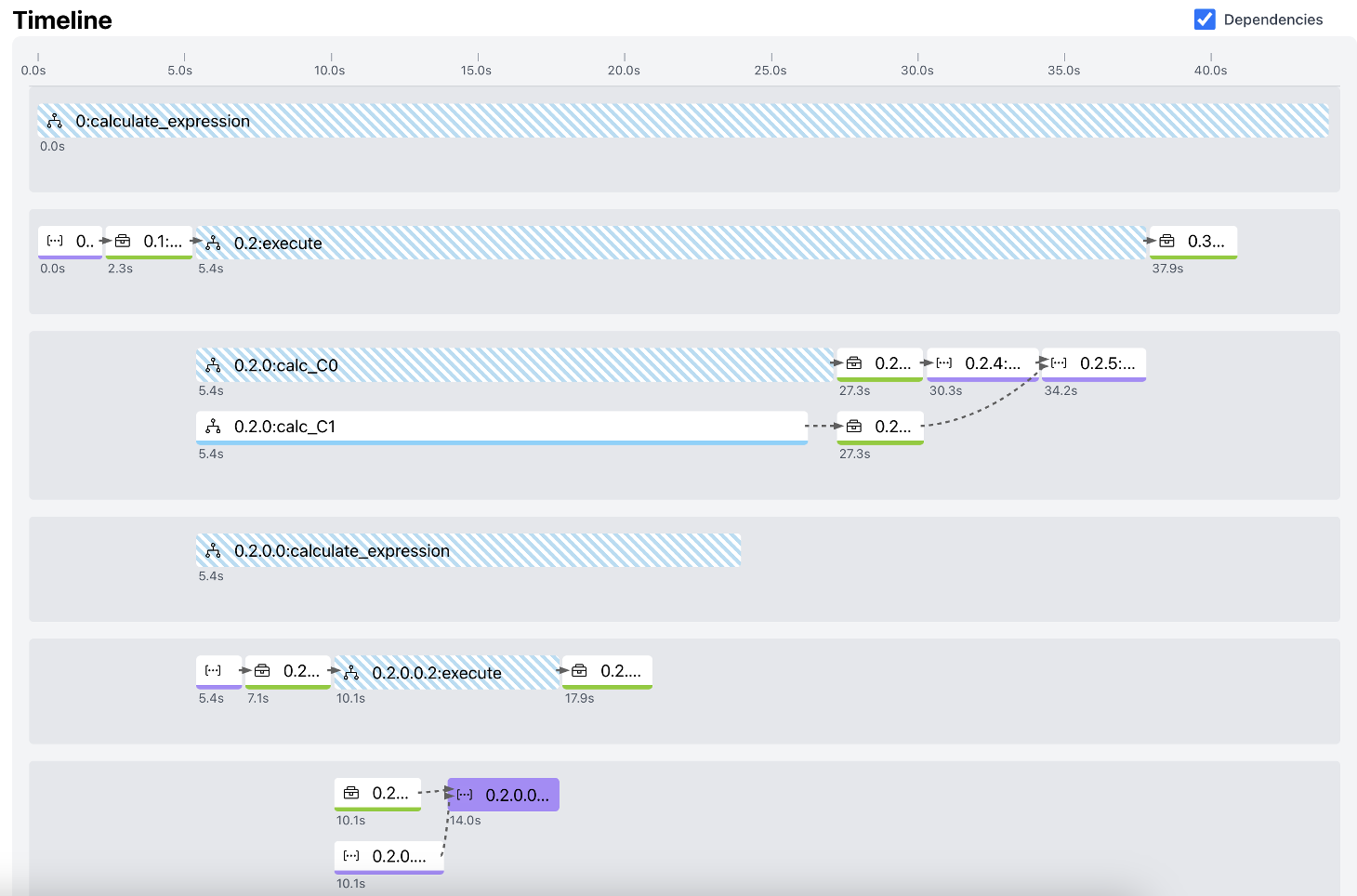} 
    \caption{Hierarchical Task Flow graph}
    \Description{A screen capture of an application showing a hierarchical task flow graph.}    
    \label{fig:calculator_example}
\end{figure}

A key example is \textbf{task flow analysis}, which represents the hierarchical breakdown of tasks into subtasks, structured as a directed acyclic graph to capture execution dependencies and lifecycle details. Metrics can be attached to each task in the hierarchy, providing insights such as LLM usage statistics and aggregated success and error rates across subtasks. This analysis provides a deeper understanding of execution patterns and is used in our benchmarking. Figure \ref{fig:calculator_example} illustrates a task flow of our calculator example.

\textbf{Exploratory analytics} enables incremental and iterative exploration, allowing users to troubleshoot agentic workflows through chat interfaces or dedicated dashboards operating on collected data.

\begin{itemize}[left=0pt]
    \item \textbf{Optimizations} fine-tune agent behavior to balance quality, performance, and cost.
    \item \textbf{Root cause analysis} identifies reasons for failures.
    \item \textbf{Comparison analysis} detects issues by comparing flows and outcomes across traces.
    \item \textbf{Retrieval-based analysis} extracts relevant information from collected data.
\end{itemize}

A key example is optimization of multi-agent systems, where SLAs define quality, latency, and cost requirements. Traditionally, one metric was prioritized for predictability, but as AI-driven multi-agent architectures evolved, their interdependencies grew more complex.  Additionally, variations in the execution paths of agentic systems lead to fluctuations in cost, latency, and quality that must be balanced to meet objectives.

\begin{figure}[H]
    \centering
    \includegraphics[width=0.8\columnwidth]{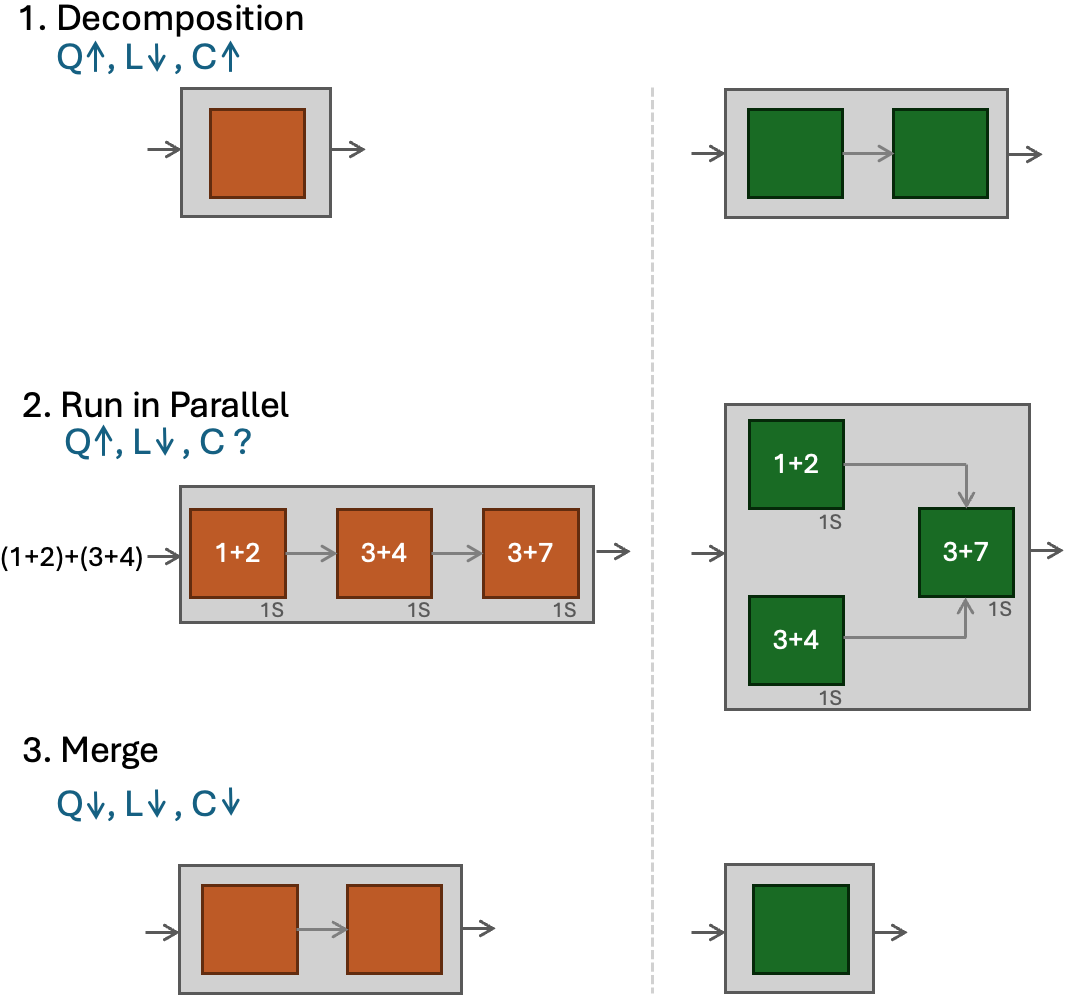}
    \caption{Optimization patterns for agentic systems.}
    \Description{A diagram showing multiple optimization patterns for agentic systems.}
    \label{fig:opt-patterns}
\end{figure}

To navigate these trade-offs, we identify key optimization patterns—decomposition, parallel execution, and merging—shown in Figure \ref{fig:opt-patterns}. While illustrative, additional strategies exist beyond the scope of this paper.

\begin{enumerate}[left=0pt]
    \item \textbf{Decomposition:} Tasks with suboptimal quality can be split into finer subtasks for greater precision. Automated detection enables adaptive restructuring, with a dedicated agent defining and assigning new tasks. This improves quality but increases cost and latency.
    \item \textbf{Parallel Execution:} Low-dependency tasks can run concurrently to reduce latency. Identifying parallelization opportunities in task dependencies minimizes execution time while preserving quality, though cost impact may vary.
    \item \textbf{Merging:} In some cases, task consolidation can improve efficiency by reducing redundant computations. Tasks that exhibit overlapping objectives or resource utilization can be merged into a more cohesive unit, improving both latency and cost.
\end{enumerate}


By leveraging an agentic framework, we propose using LLMs as system judges alongside advanced classical optimization methods to recommend improvements based on evolving SLAs. This approach refines task structures, reassignments, and execution strategies.


\textbf{Interventional analytics} actively modifies system behavior to explore and optimize performance. These changes require support from agentic middleware and explicit developer approval to ensure proper control. While some interventions can be automated, our user study shows that only 16\% of participants will trust the system to do so.

\begin{itemize}[left=0pt]
    \item \textbf{Enhanced monitoring} temporarily expands data collection
    \item \textbf{Feedback streaming} feeds analytics insights back into the agentic system, influencing real-time decision-making by agents.
    \item \textbf{Configuration tuning} dynamically adjusts system parameters.
    \item \textbf{Task injection} introduces synthetic tasks to assess impact on system behavior, performance, and adaptability.
    \item \textbf{Component replacement} allows substituting system components (e.g., LLMs, tools, or agents) with alternative or synthetic counterparts for comparative analysis.
\end{itemize}

Enhanced Monitoring collects extra runtime data but impacts performance and cost, making continuous use impractical. Instead, it activates selectively for fault detection or audits. For instance, if drift or errors exceed a threshold, the system temporarily enables detailed logging, gathering data for a set period before deactivation, ensuring efficient issue investigation without excess overhead.

This taxonomy offers a standardized framework for optimizing agentic systems, facilitating a balance between observability, adaptability, and control, and serves as a common ground for further researcher in this direction.

%% file: Sections/benchmarking.tex

%
\begin{figure*}[ht] 
    \centering
    \includegraphics[width=\textwidth]{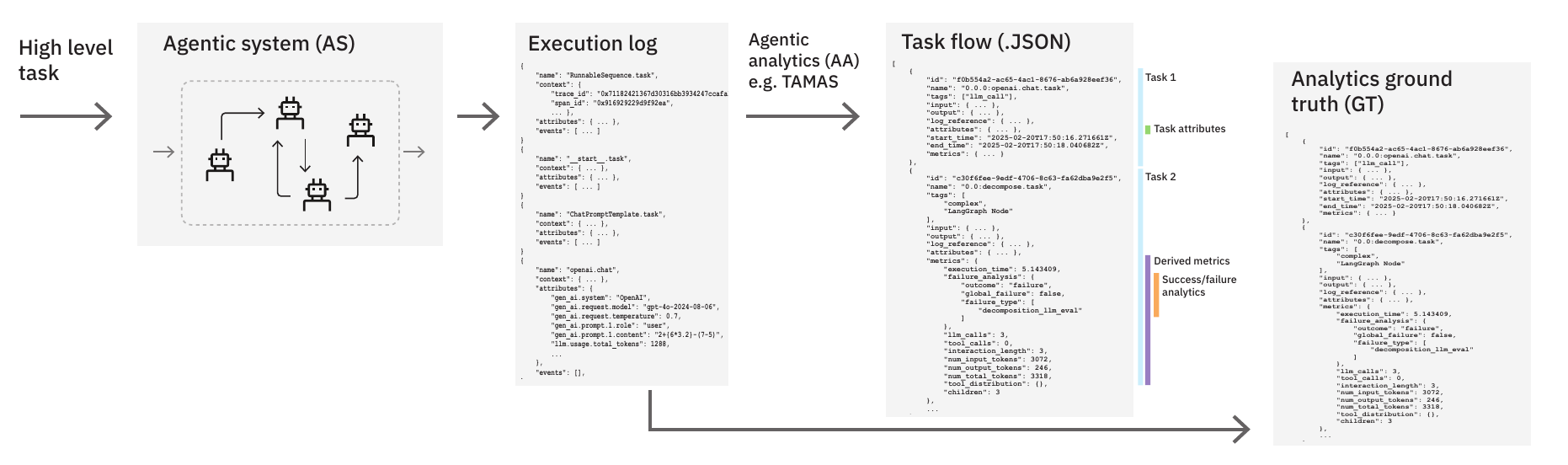} 
    \caption{Agent analytics outcomes and benchmark overview.}
    \label{fig:benchmark_overview}
\end{figure*}
While Section \ref{subsec:related_art_benchmarking} explores the current landscape of benchmarking for agentic systems evaluation, we believe agent analytics technologies should also be evaluated and benchmarked to establish an objective basis for their comparison.  


What should such benchmarks include? In {\bf agentic systems benchmarking}, a basic unit is typically a task formulated in natural language, which different systems can attempt to solve.   

In contrast, for {\bf agent analytics benchmarking}, the fundamental unit is a trace log of an agentic system's attempt to solve a task. Given such a trace log, agent analytics should determine whether it represents a correct solution ("happy path"), a correct solution with some issues, or a failure. In addition to logs, agent analytics benchmarks should include ground truth (GT) data corresponding to correct analytics outputs based on the considered logs. For each log, GT should contain at least the following components.
\begin{itemize}[left=0pt]
    \item {\bf Task flow:} A structured representation of an agentic system's trace. A task flow discovery algorithm, as discussed in Section \ref{subsec:analytics_taxonomy}, is required to extract this information.    
    \item {\bf Metrics and failure summary:} Includes key trace metrics such as execution time, number of input and output tokens, and the number of LLM and tool calls. It should also contain the number and severity of failures, both overall and categorized by failure type.
    \item {\bf Detailed failure list:} A comprehensive list of failures with descriptions.
\end{itemize}
Finally, the benchmarks can incorporate analytics outputs from other agent analytics systems to enable comparison against GT. The evaluation criteria for comparing GT with agent analytics outputs can be approached in multiple ways. One method is to compare summaries, where the simplest metric for assessing an agent analytics system versus GT is the fraction of logs with identical lists of metrics and failure counts. For failure list comparison, text similarity scores can be applied. Additionally, task flow comparisons can leverage graph-edit distance and similar metrics to assess structural differences.
An overview of worflow from initial task to agent analytics outcomes are presented in Figure \ref{fig:benchmark_overview}.

We introduce the publicly available ABBench\footnote{\url{https://github.com/genai-analytics/publications}}, which is, to the best of our knowledge the first benchmark for agent analytic systems, designed to systematically evaluate and enhance the observability of LLM-based multi-agent systems.

As a case study, we present TAMAS (Task-oriented Analytics for Multi-Agentic Systems), a proprietary agent analytics system developed at IBM. TAMAS extends standard observability by leveraging OpenTelemetry. Additionally, it incorporates advanced analytical capabilities, including task flow discovery. This system serves as a comprehensive framework for evaluating agentic workflows, identifying execution bottlenecks, and facilitating performance optimization.

The structure of ABBench is as follows:
\begin{enumerate}[left=0pt]
\item {\bf Logs:} We provide a dataset of 30 structured logs representing different agent flows, designed to capture various types of agentic system issues. These logs were generated from runs of the calculator agentic system, introduced in Section \ref{subsec:calculator}. This agentic system demonstrates many advance characteristic: distribution over multiple services, parallel processing and dynamic creation of sub-tasks. The input examples tested different calculator capabilities: 14 were purely numerical, while 16 contained natural language components. Additionally, 13 examples involved distributed calculations, and 3 inputs contained syntax errors. Out of 30 inputs, the calculator successfully solved 21 examples, while 9 resulted in incorrect answers. 7 traces constituted ``happy paths'' with no issues, while others involved some issues even when the final result has been correct. We carefully curated this dataset and plan to expand it by logs from other agentic systems in the future.
\item {\bf Ground truth analytics outcomes:} According to GT guidelines above, each log contributes to a row in the summary of metrics and failures, a task flow file and a list of failures that is empty for ``happy paths''. We categorized the following types of failures, which are characteristic of typical AS, not just the considered calculator:
\begin{itemize}[left=0pt]    %
    \item {\bf Instruction violation failures}: These occur when the decomposition of a complex mathematical expression or the planning of computational subtasks fails to fully adhere to the given instructions. More broadly, this type of failure can manifest in any LLM-based agent that disregards instructions provided in the prompt.
    \item {\bf Incorrect input failures}: Occurs when an internal calculator module receives unprocessable input, such as incorrect syntax in a mathematical expression or ambiguity in a natural language problem. This type of failure can arise in an agentic system struggling with parsing or interpreting input correctly.
    \item {\bf Validation failures}: These occur when an internal calculator computation fails the final validation check. In an agentic system, this can happen when agents produce outputs that violate correctness criteria due to accumulated errors in task execution or inconsistencies in shared state updates.
    \item {\bf Validator failures}: Occurs when correct actions of an agentic system do not pass validation checks. In the calculator setting, it can occur, for example, if properly formulated NL tasks are classified as incorrect or ambiguous ones at the validation stage.
\end{itemize}
\item {\bf TAMAS analytics outcomes:} Along GT guidelines, these outcomes also include a summary table,  task flow files and failure lists. However, TAMAS cannot detect the validator failures and, therefore, detects less failures than the Ground Truth. In addition, TAMAS does not provide correct report for logs with syntax errors in the initial input, returning empty flow files and zero metric values.
%
\end{enumerate}

\begin{table}[h]
  \begin{tabular}{p{0.41\linewidth}p{0.09\linewidth}p{0.12\linewidth}p{0.2\linewidth}}
    \toprule
Input & Output & Ground Truth & Issues  \\
\midrule
${(8-2)*3}-(5+(11/2))/5$ & 15.9 & 15.9 & ``happy path'', no issues \\
\midrule  
$[4+8*(5-3)/2]-15+(7-(9/3))$ & 1.0 & 1.0 & 1 instruction violation \\
\midrule
2+\{6*[12-(\{Multiply the sum of three, seven, and five by two. Then, subtract fifteen.\}+3)]\}/3+4*(7-5)-2/1 & None & -4.0 &
instruction violation and 3 validation failures \\
\midrule
14-\{If you subtract 3 from 43 and then divide by 5, what is the result?\} & 6.0 & 6.0 & 1 validation failure and {\bf 1 validator failure} \\
\midrule 
\{Thomas withdraws \$10000 in 20 dollar bills from the bank account.  He loses 100 bills while getting home.  After that, he uses half of the remaining bills to pay for a bill.  Thomas then triples his money.  He then converts all his bills to 10 dollar bills.  How many 5 dollar bills does he have?\} 
& None & 1200 & 6 incorrect input failures and {\bf 3 validator failures} \\
\bottomrule
\end{tabular}
\caption{Brief summaries of five benchmark cases}
\label{table:benchmark_examples}
\end{table}
Table \ref{table:benchmark_examples} summarizes the information for five benchmark cases. The first row presents the "happy path" example, where the computation was performed without any issues. 

For the next two inputs, TAMAS correctly identified the issues in the log. In the second case, the system failed to follow instructions and could not extract the upper-level brackets from the mathematical expression, although the flow was completed successfully in the end. In the third case, a similar issue occurred, and additionally, the calculator was unable to compute the following expression:
\[
6 \times [ 12 - ( \{ \text{Multiply the sum of three, seven, and five by two.} \]
\[ \text{Then, subtract fifteen} \} + 3 )] \]
The correct result is -36, but the calculator provided three different incorrect answers, which were correctly detected as validation failures.

Finally, the last two rows demonstrate cases where TAMAS failed to detect all relevant issues. In the fourth case, the correct answer for the expression inside the braces is 8. However, this correct answer was rejected by the calculator's validation module, and TAMAS did not identify the issue. As a result, we added a validator error to the ground truth. In the final case, the calculator erroneously reported that the text within the braces was not expected to return a single numerical value, leading to additional instances of validator errors that TAMAS currently cannot detect.

Overall, TAMAS summary matches GT in 60\% of benchmark cases. Other agent analytics systems are welcome to access and analyze ABBench logs and compare their outcomes with the Ground Truth.

%% file: Sections/future_work.tex
In this work, we present AABench, a benchmark designed to systematically evaluate the performance of agent analytics systems, addressing key challenges in assessing LLM-based agentic systems. Our findings highlight the inherent variability and difficulties in this evaluation process, reinforcing the need for more advanced analytical frameworks. While our proposed benchmarking approach and observability taxonomies provide a foundation for systematic evaluation, further refinements are necessary to enhance interpretability and actionability. Future efforts should focus on extending these taxonomies to capture more nuanced agent behaviors, particularly context-dependent decision-making, long-horizon dependencies and multi-agent interactions. Additionally, expanding benchmarking methodologies to incorporate dynamic evaluation scenarios will improve their applicability to real-world deployments.
As agentic systems become more prevalent, establishing standardized evaluation criteria and benchmarks across different domains and use cases will be crucial.
Beyond evaluation, significant work remains in developing practical tools and techniques for agentic system analytics and optimization, while reducing the impact of non-deterministic behaviors. This includes developing automated techniques for prompt engineering to mitigate variability, identifying execution bottlenecks through causal analysis, and optimizing agent performance by providing targeted recommendations. Further exploration into real-time performance monitoring and adaptive optimization strategies could significantly improve the robustness and efficiency of agentic systems in practical applications.

%% file: Sections/appendix.tex
\section{Datasets Used in the Experiment}

\label{sec:appendix_datasets}

All examples in the datasets involve standard mathematical operations, including parentheses, multiplication, division, addition, and subtraction.
\begin{itemize}
    \item \textbf{Pure Math:} This dataset consists of purely mathematical expressions (numbers and operations) that contain at least ten mathematical operations and include at least two nested parentheses. Example: ``10/([9*(6-5)]+8)-(3+7)''
    \item \textbf{Half NL:} This dataset includes a combination of mathematical expressions and natural language (NL) descriptions of certain expressions. To clearly distinguish between the two input types, the NL text is enclosed within curly brackets. These examples were generated by modifying the Pure Math dataset, replacing one of the numerical values with a mathematical problem whose solution corresponds to that number.
Example: ``10/([{Five added to twice the difference between twenty and the sum of seven and three.}*(6-5)]+8)-(3+7)''
    \item \textbf{Pure NL:} Pure NL: This dataset, derived from the GSM-plus dataset \cite{li2024gsm}, consists entirely of complex mathematical problems presented in natural language.
Example: ``{To participate in the local community tree-planting campaign, Mr. Julius planted twenty White Oak trees and twice as many Lodgepole Pine trees on the first day. On the second day, he planted additional White Oak trees and 1/4 more Lodgepole Pine trees than on the first day. If the total number of trees planted over both days is 140, how many more White Oak trees did Mr. Julius plant on the second day?}''
\end{itemize}


\section{Detailed Description of User study}

\label{sec:detailed_user_study}

\subsection{Survey Objectives}
This study conducted a survey to validate the arguments and con-
cepts presented in this paper, examining their applicability across
the real-world lifecycle of agentic system (AS) management, includ-
ing development, testing, and deployment.

The survey focused on three primary research areas:
\begin{itemize}
   \item \textbf{Characteristics and usage patterns of agentic systems}: The first section of the survey gathered insights into respondents’ backgrounds, experience levels, development
    environments, and preferred tools. This data provided a
    foundation for analyzing whether differences in perspec-
    tives stemmed from variations in background or usage pat-
    terns.
   \item \textbf{New Challenges:} Assess practitioners' perceptions of the challenges inherent in agentic systems, particularly concerning non-deterministic behaviors, the need for deeper runtime insights, and the balance between quality, latency, and cost.
   \item \textbf{Tooling Gaps:} Assess the effectiveness of current analytics tools and methodologies in addressing the emerging challenges of the LLM era, while identifying requirements and gaps in development, testing, and maintenance tools
\end{itemize}

\subsection{Methodology}

\subsubsection{Participants} \hfill

\textbf{Target population}
The survey targeted professionals, predominantly those affiliated with research departments. Practitioners possess experience in the development, testing, and deployment of agentic systems. The cohort represented a spectrum of expertise, ranging from individuals with foundational experience to those with extensive, advanced expertise in agentic systems. Participants came from diverse backgrounds with respect to the frameworks they utilized for building these systems, as well as the observability and analytical tools they employed to monitor and evaluate system performance.

\textbf{Number of participants} A total of 38 participants were recruited for the survey

\textbf{Recruitment} Recruitment was conducted by reaching out to diverse teams, with the goal of collecting feedback from a wide spectrum of practitioners. Efforts were made to include participants representing various use cases and domains, such as prompt researchers, evaluation analysts, developers, engineers, and other relevant roles

\subsubsection{Survey structure}\hfill
The survey was structured from 4 sections
\begin{itemize}
   \item \textbf{General and technical background:} This section consisted of 6 multiple-selection questions covering the practitioner’s level of experience, their role (e.g., development, testing, or production), as well as the frameworks, LLMs, and tools they commonly use.
   \item \textbf{Requirements for observability and analytics tools:}  In this section, participants answered a series of 5 multiple-selection questions, allowing them to share details about the nature of their work with agentic systems, their expectations for analytics tools, and the features that could enhance their workflow.
   \item \textbf{Likert-scale questionnaire:} A section in which participants were presented with a list of 9 statements and asked to indicate the extent of their agreement with each statement. The statements focused on participants' experiences, challenges, and requirements related to agentic systems. 
   \item\textbf{Two open-ended questions were included} the first asked participants to identify the primary pain points they face when dealing with agentic systems, while the second provided an opportunity to share any general comments about this domain.
\end{itemize}

\textbf{Data Collection Procedure} The data for the survey was collected through an online, unmoderated format using a widely adopted surveying platform. Participants were free to begin the survey at their convenience and were not constrained by any time limits, allowing them to complete it at their own pace. None of the questions were mandatory. Prior to participation, all respondents provided informed consent for their data to be retained and used for research purposes. The survey was conducted during January 2025.

\subsection{Survey results}
\subsubsection{Quantitative Data:}
As previously noted, the survey's objectives targeted three primary areas. The first area focused on characterizing the landscape of agentic systems practitioners, including their preferred frameworks, tools, and expectations. To capture this information, the following questions were posed, yielding key data points:

\begin{itemize}
   \item \textbf{Question: My experience in the Gen AI domain?} (\autoref{fig:Domain experience})
   With 92\% of respondents identifying as intermediate or expert, the cohort is composed of highly experienced practitioners, strengthening the credibility of their insights  
\begin{figure}[H]
  \centering
  \includegraphics[width=\linewidth]{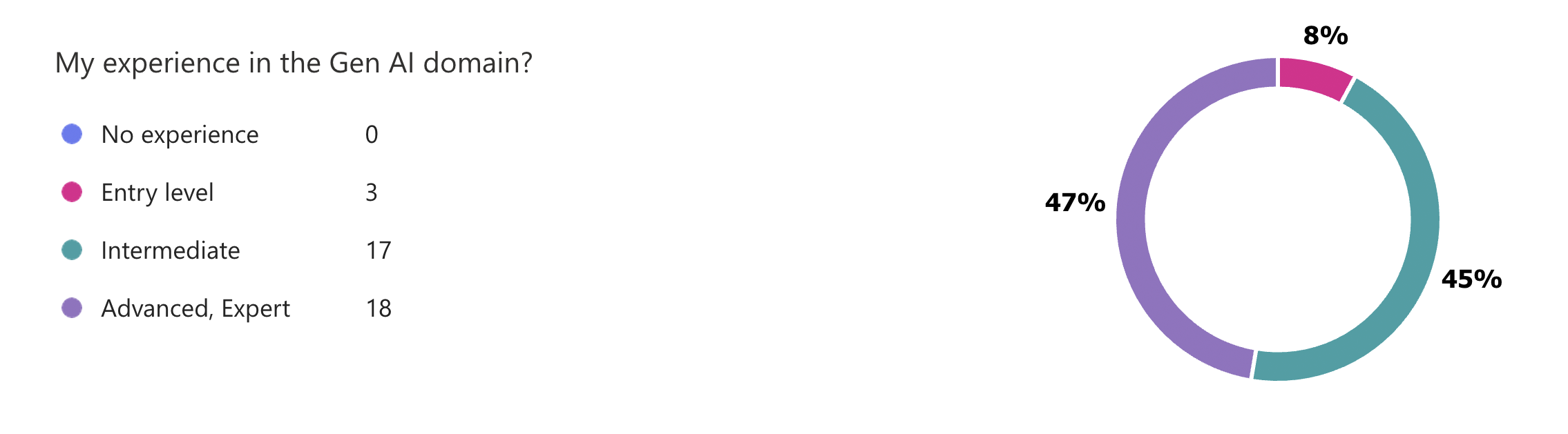}
  \caption{My experience in the Gen AI domain?}  
  \Description{A diagram showing the survey results with respect to the question: My experience in the Gen AI domain.}  
  \label{fig:Domain experience}
\end{figure}

   \item \textbf{Question: Agentic Framework(s) I use} (\autoref{fig:Agentic frameworks I use})
   The respondents primarily use widely adopted frameworks - LangGraph, LangChain and CrewAI, which reflects the prevailing standards in agentic system development today and lends relevance to their perspectives and responses
\begin{figure}[H]
  \centering
  \includegraphics[width=\linewidth]{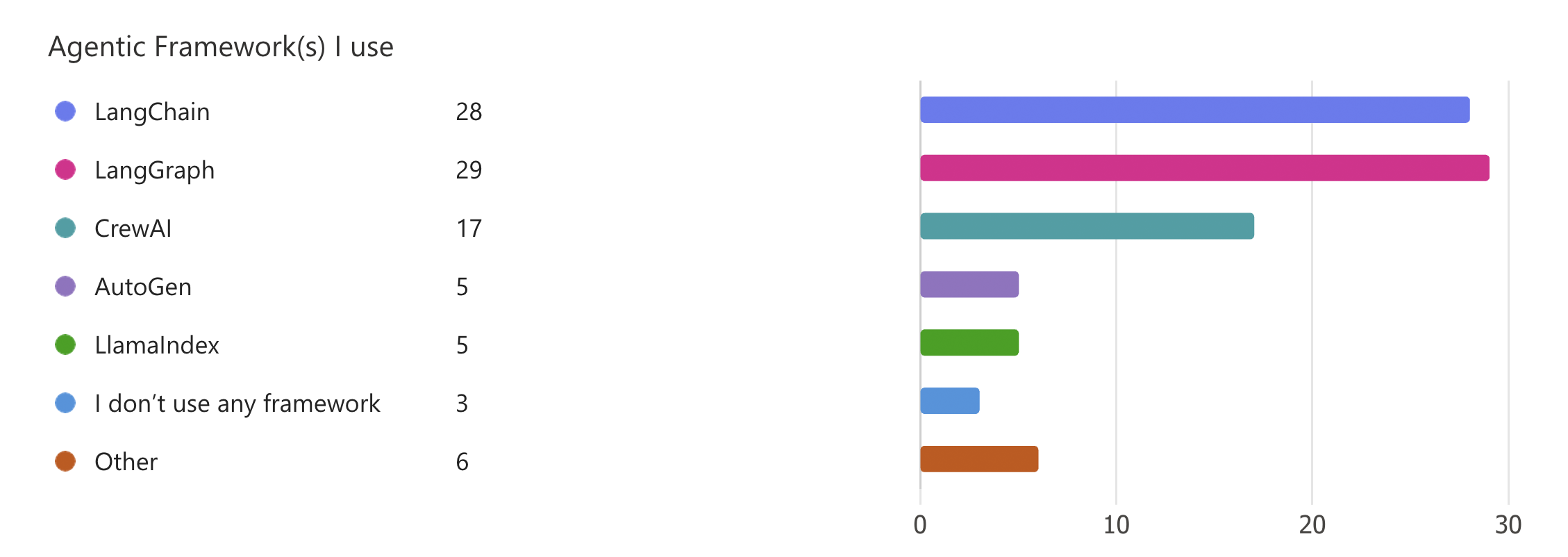}
  \caption{Agentic frameworks I use}  
  \Description{A diagram showing the survey results with respect to the question: Agentic frameworks I use.}  
  \label{fig:Agentic frameworks I use}
\end{figure}

   \item \textbf{Question: Current tool I use or used to observe \& explore agentic systems.} (\autoref{fig:Current observation tools})
   Respondents report using a variety of tools, yet they encounter similar problems and missing features across them. This suggests that the gaps in tool capabilities are widespread and not limited to any specific tool.
\begin{figure}[H]
  \centering
  \includegraphics[width=\linewidth]{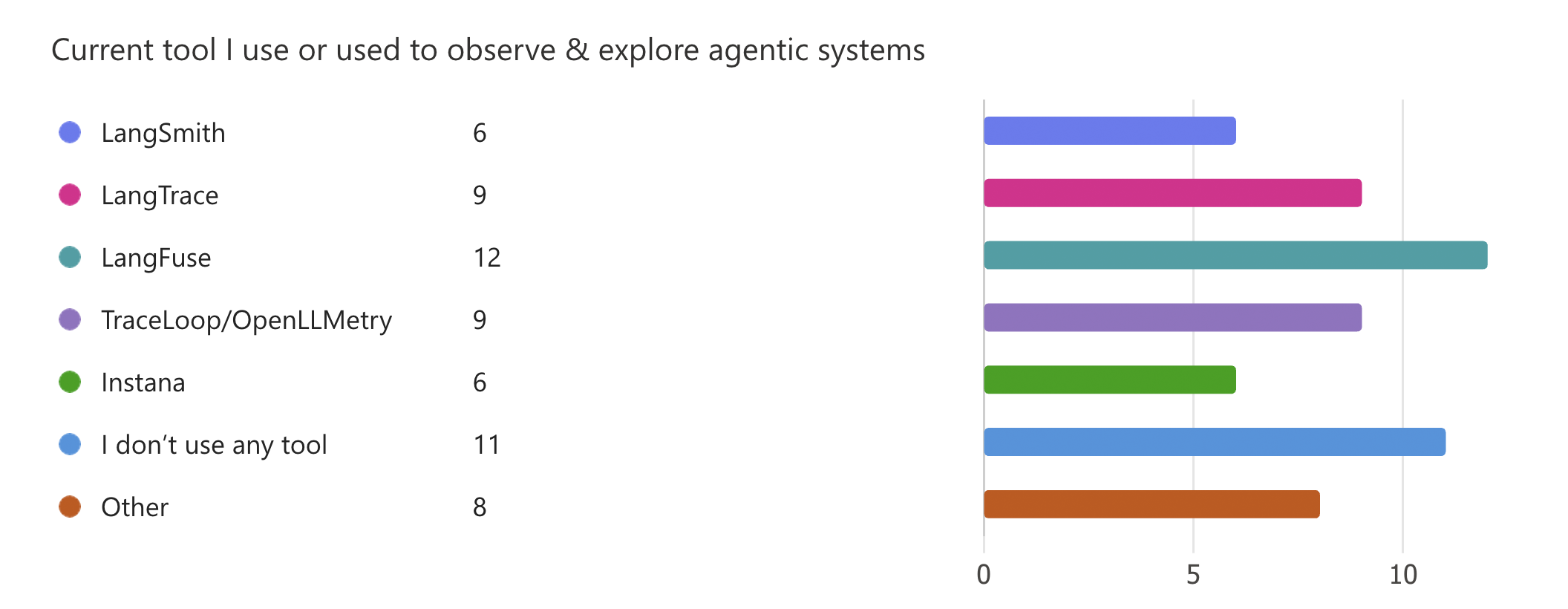}
  \caption{Current observation tools}  
  \Description{A diagram showing the survey results with respect to the question: Current observation tools.}  
  \label{fig:Current observation tools}
\end{figure}

   \item \textbf{Question: My primary objective for using an agent analytics tool.} (\autoref{fig:Primary objective})
  The majority of respondents - 73\% are developers, while only a few are engaged in more advanced stages, such as testing or maintaining agentic systems in production. It’s reflects the early adoption phase of the domain. 
\begin{figure}[H]
  \centering
  \includegraphics[width=\linewidth]{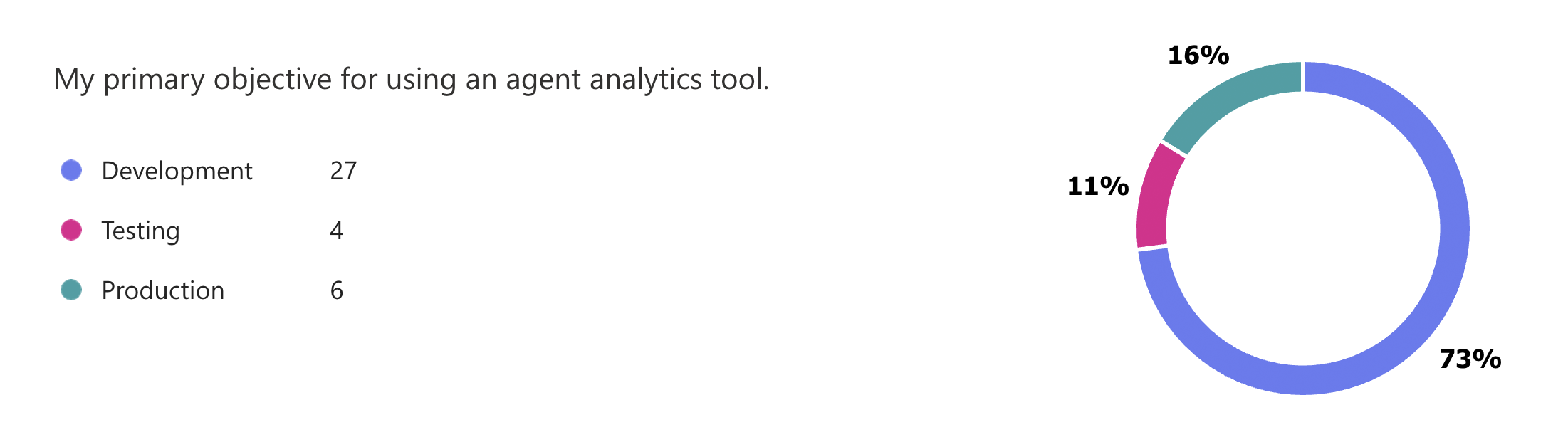}
  \caption{My primary objective for using an agent analytics tool}  
  \Description{A diagram showing the survey results with respect to the question: My primary objective for using an agent analytics tool.}  
  \label{fig:Primary objective}
\end{figure}

   \item \textbf{Question: What characteristics does my agentic system have or require?} (\autoref{fig:System characteristics})
  A significant portion of respondents' selections highlights emerging system characteristics that are particularly important in newer agentic systems. Features such as hierarchical tasks, recursive calls, parallel flows, and long-term memory management necessitating enhanced focus and advanced observability tools. 
\begin{figure}[H]
  \centering
  \includegraphics[width=\linewidth]{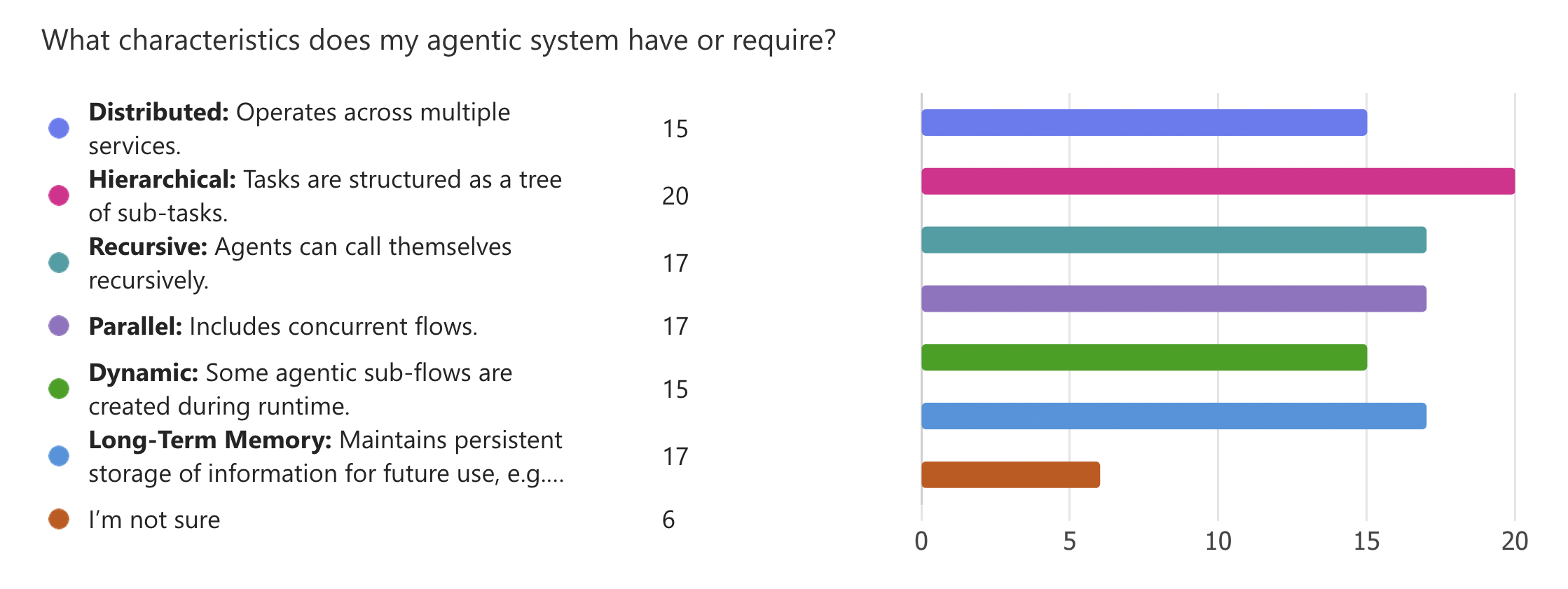}
  \caption{What characteristics does my agentic system have or require?}  
  \Description{A diagram showing the survey results with respect to the question: What characteristics does my agentic system have or require?}  
  \label{fig:System characteristics}
\end{figure}

   \item \textbf{Question: The top runtime aspects of agentic systems that I would like to investigate.} (\autoref{fig:Runtime aspects})
  These responses highlight the key aspects of agentic systems that draw the most attention in this evolving domain, such as LLM interactions and multi-agent coordination, underscoring new requirements that go beyond the capabilities of traditional observability tools. 
\begin{figure}[H]
  \centering
  \includegraphics[width=\linewidth]{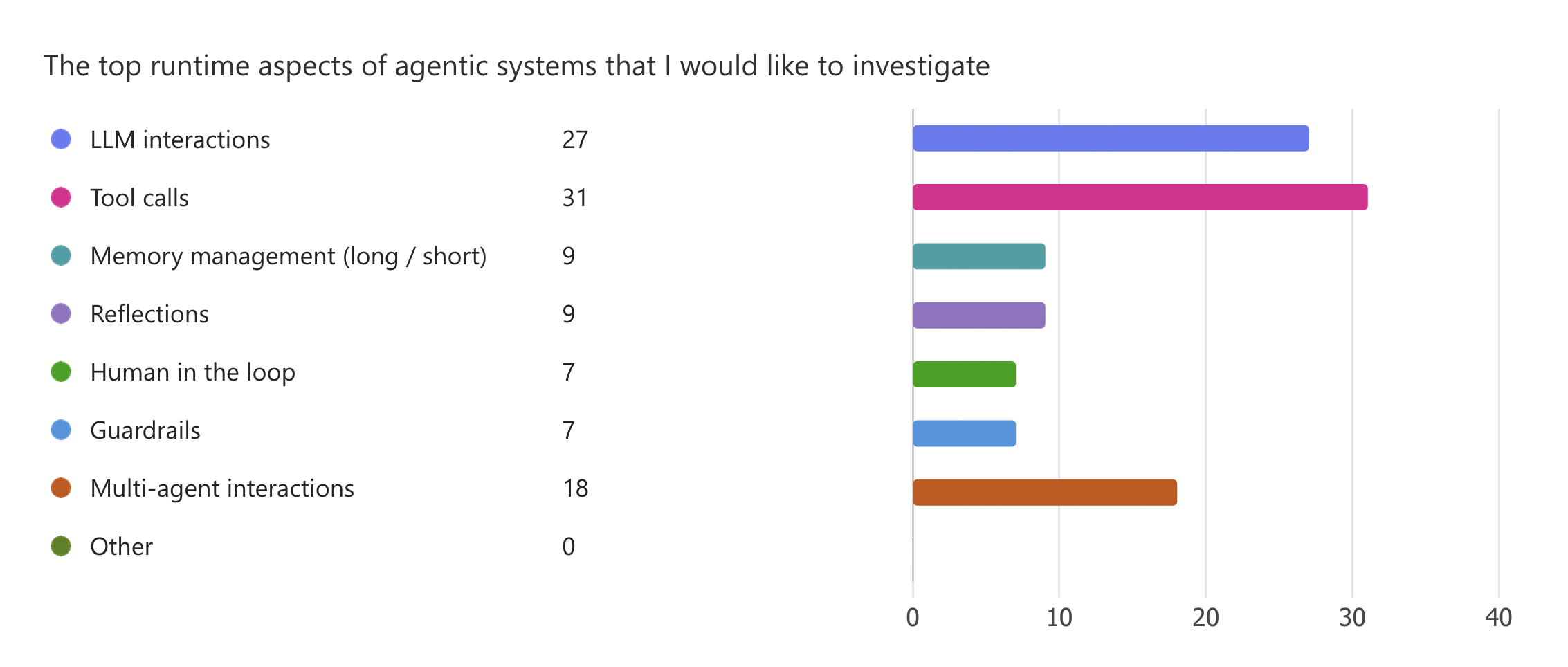}
  \caption{Runtime aspects to investigate}  
  \Description{A diagram showing the survey results with respect to the question: Runtime aspects to investigate}  
  \label{fig:Runtime aspects}
\end{figure}
\end{itemize}
The following 4 questions further illustrate the evolving requirements for enhanced observability and analytics tools in modern agentic systems, compared to those in previous generations of LLM-based systems. These requirements consistently ranked among the top priorities in nearly every aggregated response

\begin{itemize}
    \item \textbf{Question: The top types of analytics I will most benefit from.} (\autoref{fig:Analytics to benefit from})
    \begin{figure}[H]
  \centering
  \includegraphics[width=\linewidth]{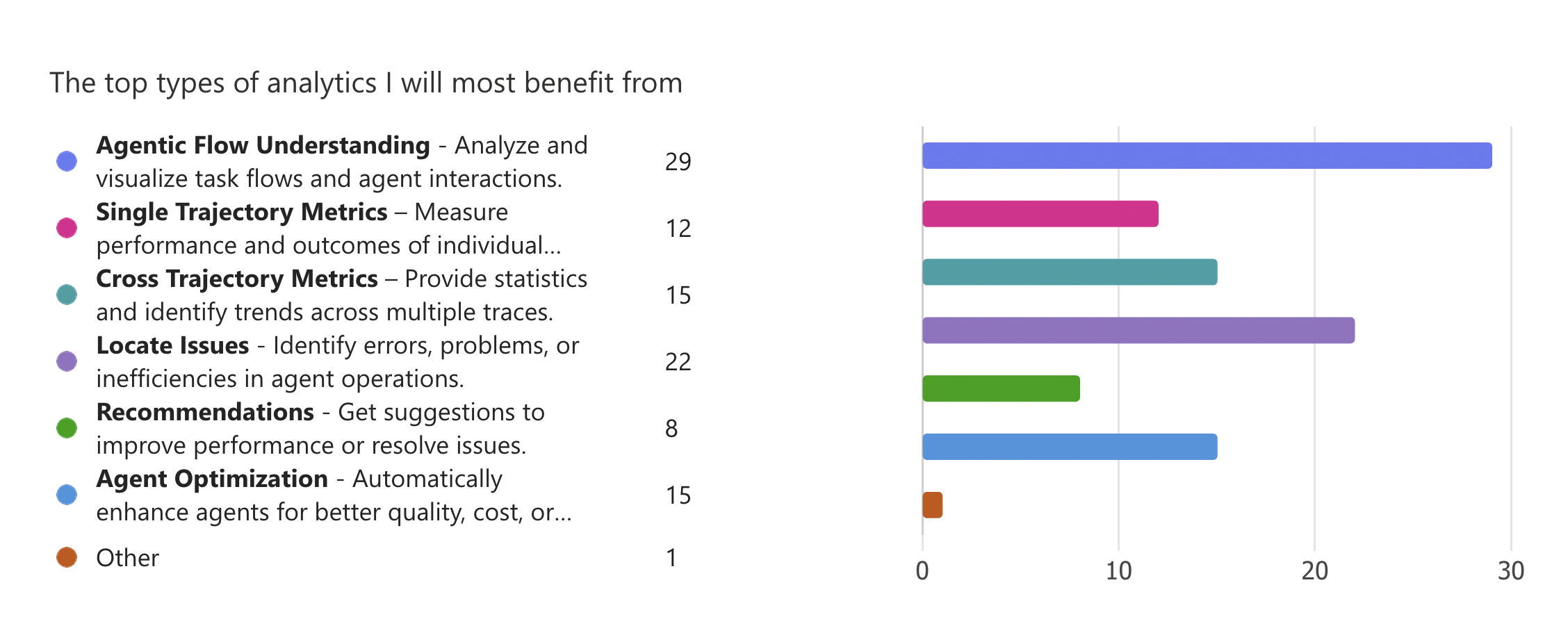}
  \caption{The top types of analytics I will most benefit from}  
  \Description{A diagram showing the survey results with respect to the question: The top types of analytics I will most benefit from}  
  \label{fig:Analytics to benefit from}
\end{figure}

    \item \textbf{Question: Which top issues would you most want help with identifying in your system?} (\autoref{fig:Issues to identify})
    \begin{figure}[H]
  \centering
  \includegraphics[width=\linewidth]{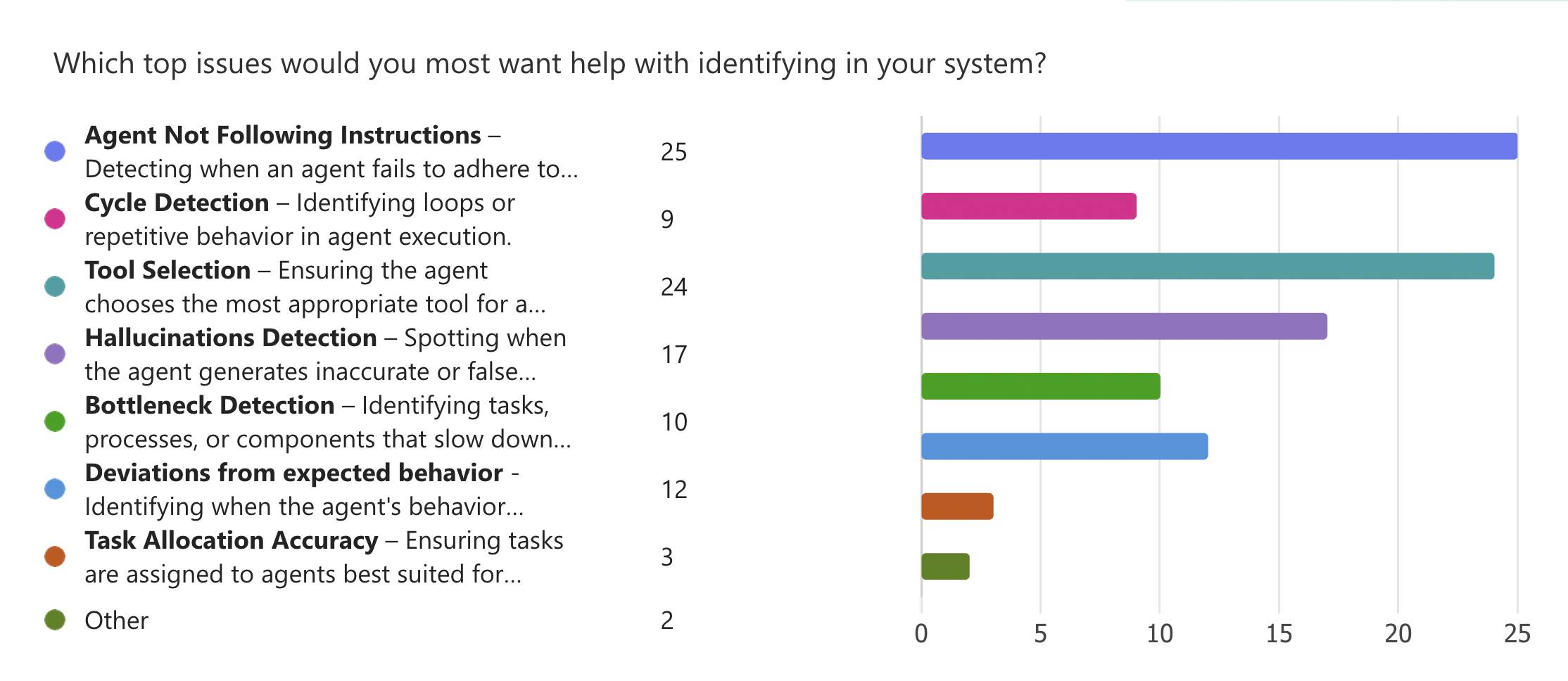}
  \caption{Which top issues would you most want help with identifying in your system?}  
  \Description{A diagram showing the survey results with respect to the question: Which top issues would you most want help with identifying in your system}  
  \label{fig:Issues to identify}
\end{figure}

    \item \textbf{Question: Select the optimizations that would benefit you the most} (\autoref{fig:Select the optimizations that would benefit you the most})
    \begin{figure}[H]
  \centering
  \includegraphics[width=\linewidth]{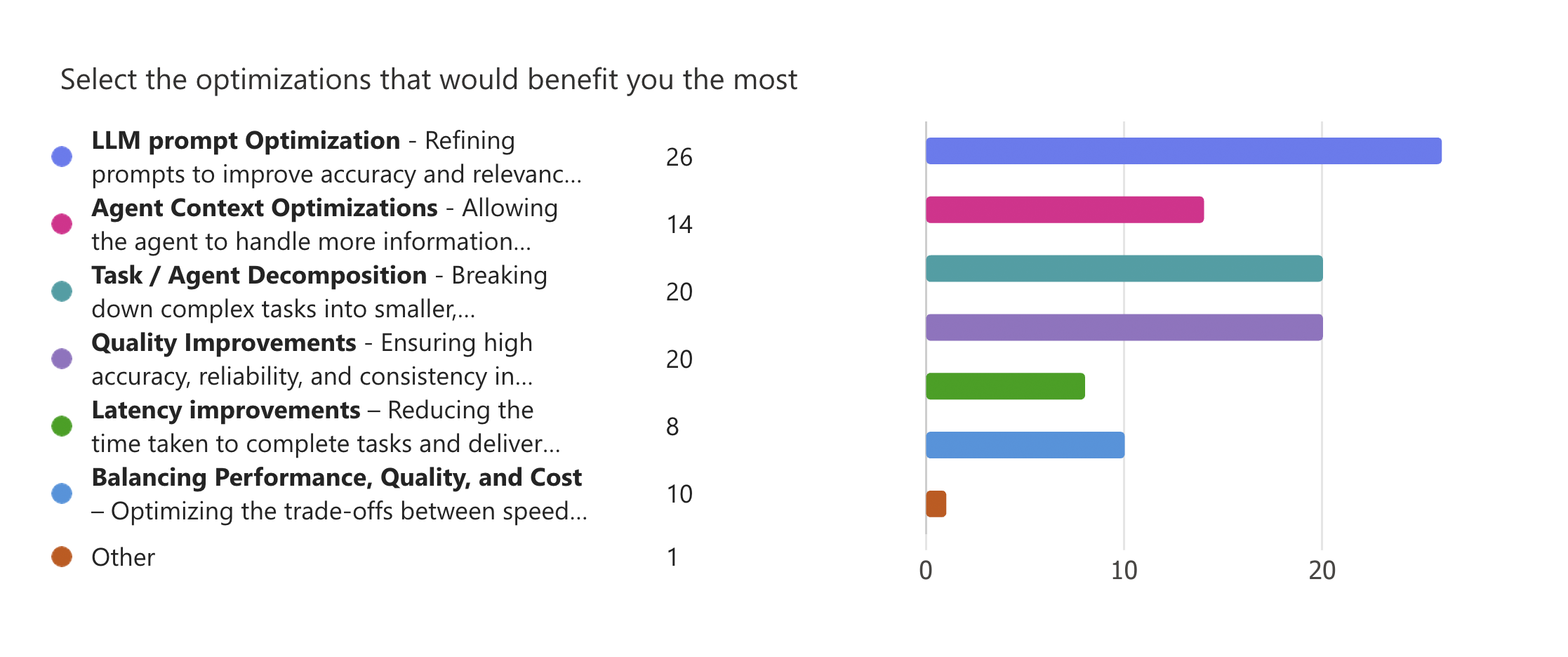}
  \caption{Select the optimizations that would benefit you the most}  
  \Description{A diagram showing the survey results with respect to the question: Select the optimizations that would benefit you the most}  
  \label{fig:Select the optimizations that would benefit you the most}
\end{figure}

    \item \textbf{Question: Choose the top aspects of Human-in-the-Loop interactions whose runtime behavior you would like to analyze} (\autoref{fig:Human in the loop})
    \begin{figure}[H]
  \centering
  \includegraphics[width=\linewidth]{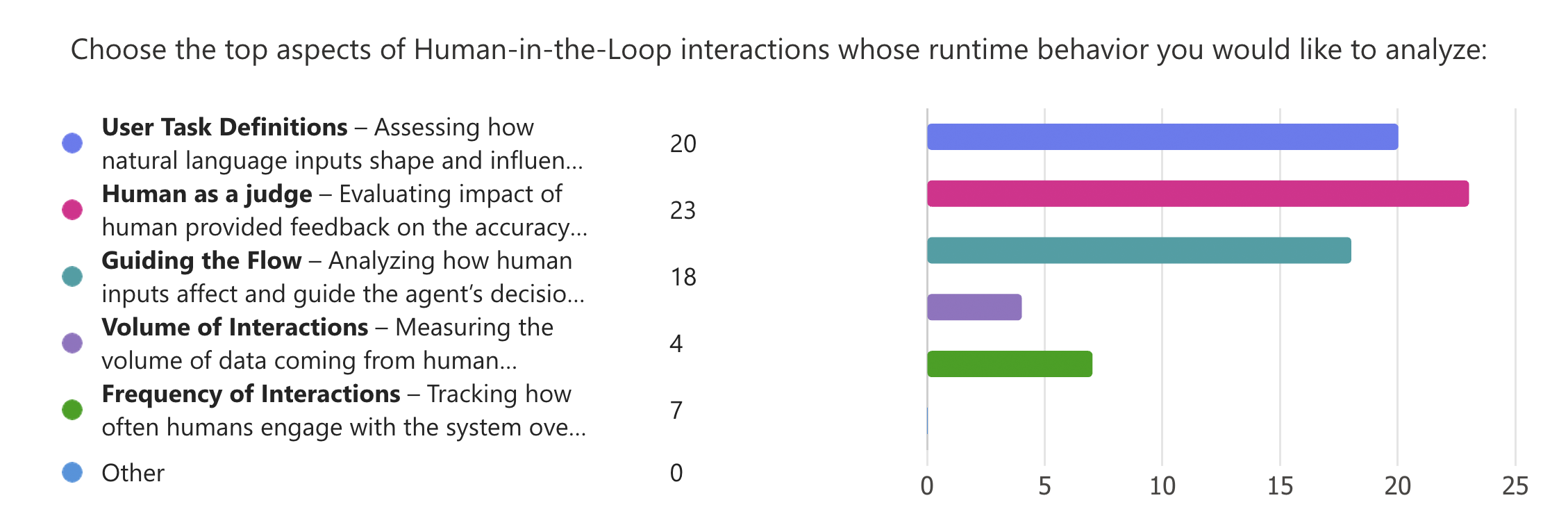}
  \caption{Human in the loop analysis}  
  \Description{A diagram showing the survey results with respect to the question: Human in the loop analysis}  
  \label{fig:Human in the loop}
\end{figure}
    
\end{itemize}

The following 4 statements, taken from the Likert-style section of the survey, further reflect the challenges practitioners face in this domain. These difficulties stem not only from the increasingly complex structure of agentic system flows but also from the limitations of the tools available to support them.

\begin{itemize}
    \item 'I want to understand the internal runtime behavior of an agentic system' (statement 2)
    \item 'The analytics tools currently available meet my needs' (statement 4)
    \item 'I find it challenging to identify the root cause of issues in agentic systems' (statement 9)
    \item 'The Non-deterministic flow of agentic systems is a major challenge' (statement 1)
\end{itemize}

 \begin{figure}[H]
  \centering
  \includegraphics[width=\linewidth]{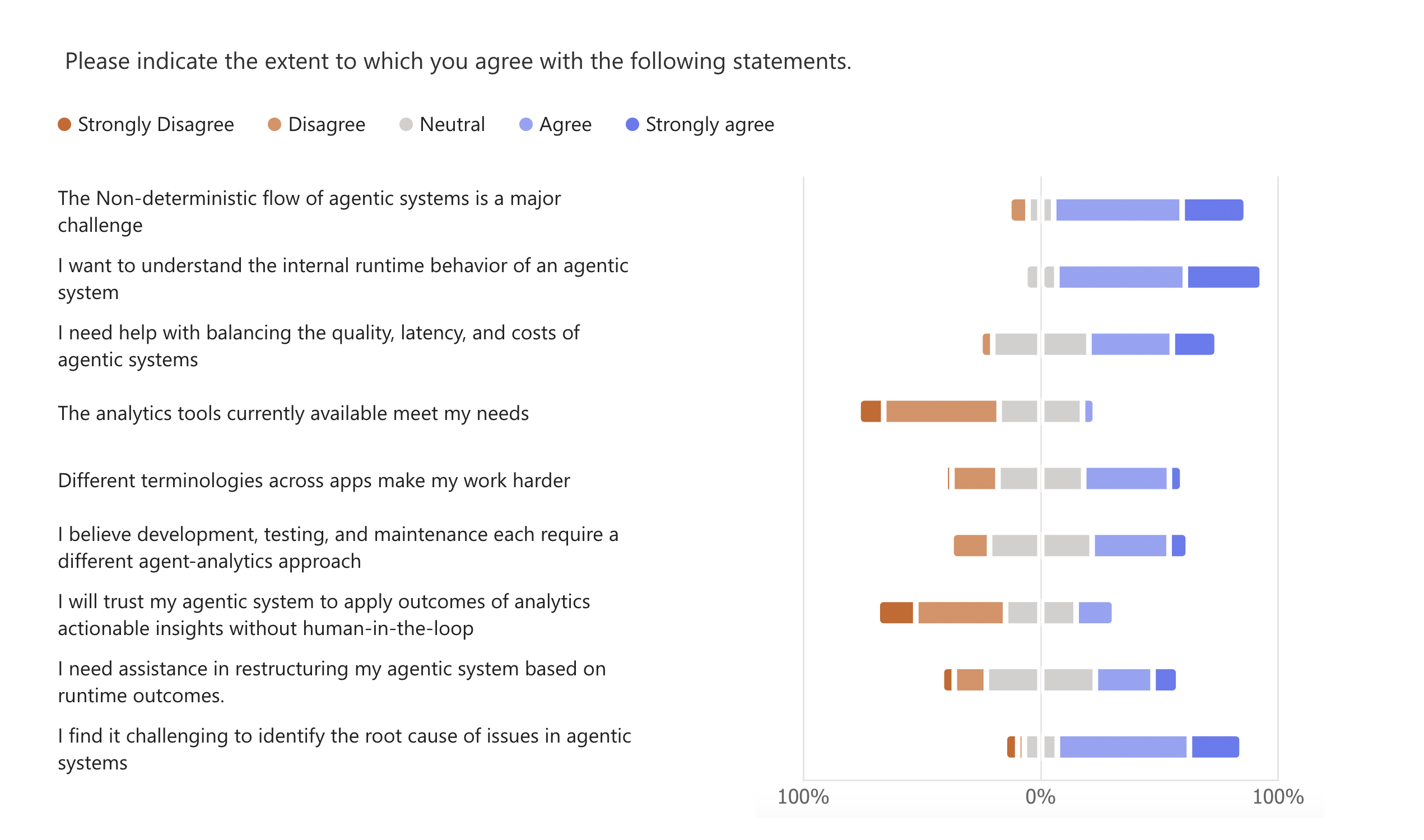}
  \caption{Domain related statements}  
  \Description{A diagram showing the survey results with respect to the question: Domain related statements} 
  \label{fig:Statements}
\end{figure}

\subsubsection{Qualitative Data:}
To gain deeper insight into the challenges faced by practitioners developing agentic systems, the survey included an open-ended question:'What is your main pain point in developing an agentic system that you believe could be addressed by analyzing its runtime behavior?'

This single question captured the diverse experiences, concerns, and priorities of agentic system practitioners. The breadth and depth of the answers highlight the complexity of agentic system development, touching on both technical and strategic concerns.

Upon analysis, responses were clustered into 5 overarching themes:

\begin{enumerate}
    \item \textbf{Debugging, Execution Understanding, and Observability Gaps:} 
    Practitioners report challenges in understanding execution behavior, debugging failures, and identifying patterns across runs. Additionally, the sheer volume of logged data makes extracting insights difficult, and existing observability tools lack support for agentic systems. Some key statements:
    \begin{itemize}
        \item 'Understanding the actual execution behavior, detecting deviations.'
        \item 'The main pain point in development of our agentic system is identifying the needed improvements. After running a test data (size=K), we need to take each trajectory (with its metadata) and analyze the reason it failed, identify patterns (across trajectories) of failures to make sure we solve cardinal problems without creating new ones.'
        \item 'The challenges in understanding the dynamic flow of the agentic workload.'
        \item 'There is too much information that is logged by different agentic frameworks and it is hard to process them and get insights.'
    \end{itemize}

        \item \textbf{Trust, Transparency, and Governance:} 
    A key challenge in developing agentic systems is ensuring trust and transparency for both users and developers. Practitioners emphasize the need for clear system behavior, governance mechanisms, and reliability in agent decision-making. Some key statements:
    \begin{itemize}
        \item 'Enable the agentic system can be transparent to customers so as to build a trusted agentic system for clients.'
        \item 'Ensure the agentic solutions' robust adaptability while maintaining trust and efficiency in production scenarios.'
        \item 'I want to be able to fully 'trust' that the output is not fluff and is accurate and up-to-date. I want to potentially, blindly trust the agentic system.'
    \end{itemize}

        \item \textbf{Prompt Adherence, Response Control, and LLM Behavior:} 
    A recurring concern is the unpredictability of agentic system responses, especially in structured outputs, tool usage, and prompt adherence. Developers struggle to enforce deterministic behavior despite carefully crafted instructions. Some key statements:
    \begin{itemize}
        \item 'Make the LLM follow the prompt."
        \item "The main point is the unexpected results we get from the LLM, in particular - tool calls. To me, the most important ability is to analyze what the LLM performs given the prompt it is provided with. Why doesn't it call the tool it should be calling? Why doesn't it return the right answer given the system prompt it was instructed with?'
        \item 'I'm not referring to nondeterministic responses, but even the format of the responses can vary, despite explicitly requesting JSON output and providing a few-shot prompt.'
    \end{itemize}

            \item \textbf{Optimization of Multi-Agent and Complex Systems:} 
    Optimizing agentic workflows, especially in multi-agent environments, is a challenge. Practitioners need better insights into how inputs influence outputs, how agents interact, and how planning structures should evolve. Some key statements:
    \begin{itemize}
        \item 'I think it's really hard to understand how to optimize multi-agent systems to generate the output I'd like. Prompts can really change their behavior and being able to produce a high-quality, semi-deterministic response is of prime importance.'
        \item 'The effect that different inputs have on the outcome of individual tasks or nodes in a large scale.'
    \end{itemize}

    \item \textbf{Rapidly Evolving Tech Landscape and Tooling Choices:} 
    The agentic systems field is evolving at a rapid pace, making it difficult for developers to choose the right frameworks and tech stacks. There is uncertainty about which tools will be sustainable long-term. A key statement:
    \begin{itemize}
        \item 'This domain evolves so quickly, it is very hard to choose the right framework, tech stack to work with.'
    \end{itemize}
    
\end{enumerate}

\subsection{Two-dimensional Analysis}

\label{subsec:two_dim}

We performed correlation analysis in order to find significant two-dimensional relations between answers to different questions of the survey. 

The following 65 features were analyzed:
\begin{itemize}
    \item A single ordinal {\em experience} feature with $(-1,0,1)$ values, where $-1$ corresponds to ``Entry level'', 0 to ``Intermediate'' and 1 to ``Advanced''.
    \item Three binary features that correspond to the primary objective for using an agent analytics tool: development, testing and production.
    \item 15 binary features that correspond to the use of different tools by survey respondents: 6 agentic frameworks, 4 observation frameworks and 5 monitoring tools.
    \item 9 ordinal features that correspond to answers to Likert-style questionnaire. Answers (Strongly Disagree, Disagree, Neutral, Agree, Strongly Agree) were transformed into $(-2,-1,0,1,2)$ scale. If a user did not provide answer to some question, it was replaced by ``Neutral''.
    \item 37 binary features related to multiple-choice questions on requirements: 6 characteristics of AS that the users require, 7 runtime aspects of AS the users would like to investigate, 6 types of analytics the users will benefit from, 7 issues the users would like to identify, 6 types of AS optimization the users would benefit from, 5 types of human-in-the-loop interactions that the users would like to analyze.
\end{itemize}
We computed the Spearman's rank correlations for 65 above-mentioned features. 

\begin{figure}[H]
  \centering
  \includegraphics[width=\linewidth]{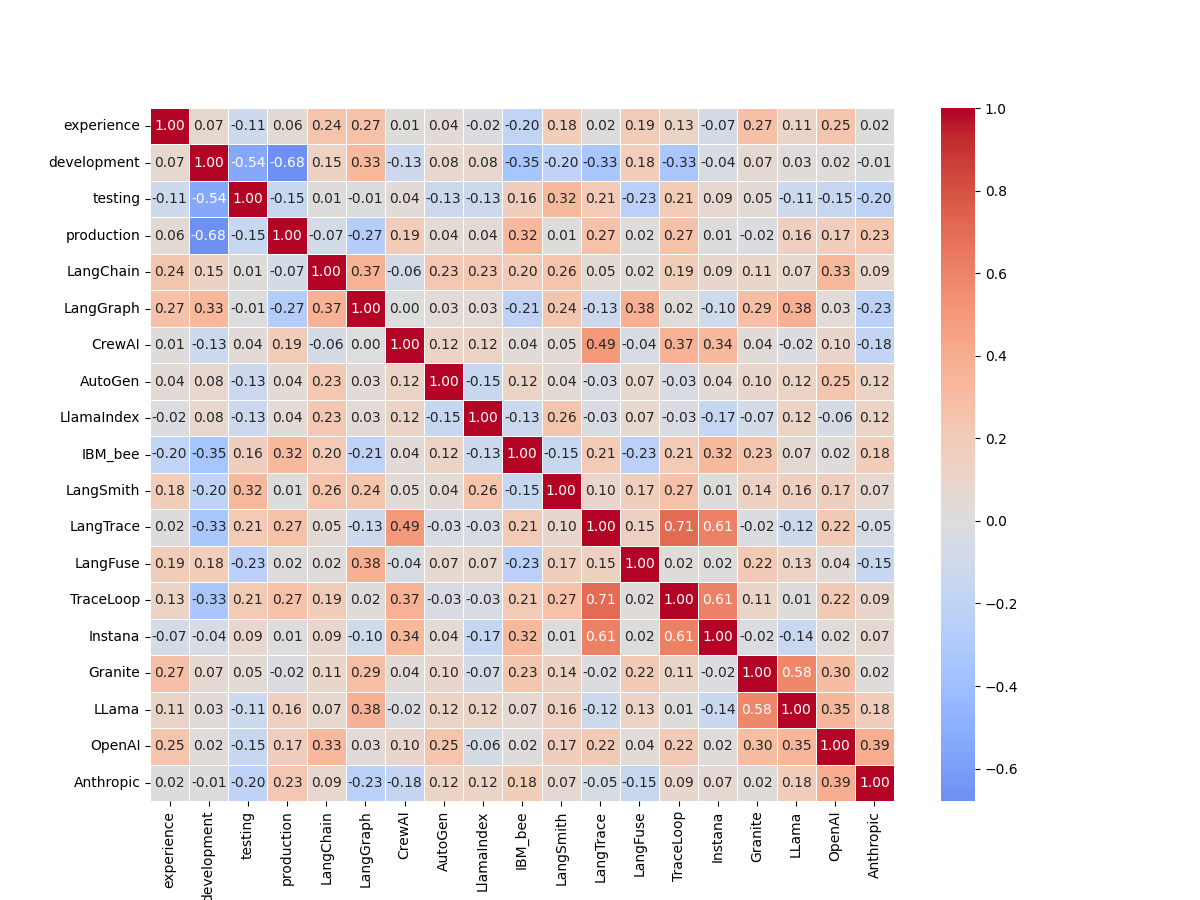}
  \caption{Correlations between experience, primary objective and tools features}  
  \Description{A diagram showing the survey results with respect to the survey: Correlations between experience, primary objective and tools features} 
  \label{fig:tool_correlations}
\end{figure}

Figure \ref{fig:tool_correlations} 
presents correlations for some of the features, described above.

Since the number of correlation matrix entries is larger than 2,000, we encounter multiple comparison problem and cannot check significance of $p$-values one-by-one. Using Benjamini–Hochberg procedure to control the False Discovery Rate (FDR) and using $\alpha=0.1$ constraint for FDR, we discover significant correlations between features.



\begin{table}[H]
  \caption{Highly significant correlations}
  \label{table:correlation}
  \begin{tabular}{p{0.38\linewidth}p{0.38\linewidth}p{0.08\linewidth}}
    \toprule
    Feature 1 &  Feature 2 & Corr. \\
    \midrule    
    use TraceLoop & use LangTrace & 0.71  \\
    \midrule
    use Instana & use  LangTrace &  0.61  \\
    \midrule
    use TraceLoop & use Instana &  0.61  \\
    \midrule
    %
    need support: IBM Granite  & need support: Llama &  0.58 \\
    \midrule
    runtime aspects to investigate: LLM interactions  & runtime aspects to investigate: tool calls &  0.58 \\  
    \midrule
    use Bee agent framework & interest in balancing latency, quality and cost &  0.57  \\    
    \midrule
    agentic systems requirements: distributed  & agentic systems requirements: dynamic &  0.56 \\   
    \midrule
    use LangFuse & top runtime aspect: human-in-the-loop &  0.55  \\
    \midrule
    \bottomrule
\end{tabular}
\end{table}

Table \ref{table:correlation} presents pairs of variables that were found significant by Benjamini–Hochberg procedure. The first three rows demonstrate the joint use of three tools: Instana, Traceloop and LangTrace. The fourth row indicates joint interest in two open-source and relatively small LLMs (versus big commercial model, such as OpenAI and Antropic). 

The fifth row shows strong positive relation between the two most popular answers to the question on top runtime aspects: LLM interactions and tool calls. In the next row we observe that the users  if Bee Agent Framework, that has been open-sourced by IBM Research are especially interested in optimization and balancing  of the main AS characteristics. Row 7 contains an interesting insight about the joint interest in AS that are distributed across multiple service and whose structure is dynamically updated during runtime. Finally, the last row indicates interest of LangFuse users to the human-in-the-loop runtime aspect; probably, LangFuse lacks this capabilities.

\subsection{Cluster Analysis}

We performed $K$-means clustering on 65 features above using dynamics of WCSS (Within-Cluster Sum of Squares) as the criterion to select the number of clusters $K$. $K=2$ has been selected. Below we summarize the main characteristic of the two clusters and differences between them.
 
Cluster 1 includes 27 participants, Cluster 2 includes 11 participants, there has not been significant difference in experience or primary objective for tool usage for the two clusters.  

If we consider 15 binary features that are related to tool use and 37 binary features that correspond to questions on requirements, relatively large differences are observed for many of them: Cluster 2 members use more tools and select more requirements. The following binary features show average differences that are larger than 0.3 with more frequent selections by Cluster 2:
\begin{description}
\item[Agentic Framework use:] LlamaIndex;
\item[LLM model use:] OpenAI, Antropic;
\item[AS characteristics:] parallel, recursive, dynamic, maintaining log-term memory; 
\item[Top types of analytics:] agent optimization;
\item[Top issue to identify:] cycle detection, deviations from expected behavior;
\item[Optimizations that would benefit most:] LLM prompt optimization, task/agent decomposition into smaller manageable subtasks, 
\item[Types of human-in-the-loop interactions:] human as a judge.
\end{description}
On the other hand, there are no features with more frequent selections by Cluster 1 if we use the same 0.3 threshold.

Now consider Likert-style questions that are coded in (-2, \ldots, 2) and select 0.5 threshold for relatively large differences. Cluster 2 member agree with the following statements more than Cluster 1 members.  
\begin{itemize}
\item I want to understand the internal runtime behavior of an agentic system;
\item Different terminologies across apps make my work harder;
\item I need assistance in restructuring my agentic system based on runtime outcomes.
\end{itemize}
Summarizing, respondents from Cluster 2 have more diversified and advanced requirements from agentic systems. Since they work with diverse applications and models, they often struggle with different terminologies across apps.


